%% file: arxiv.tex
\documentclass[10pt,twocolumn,letterpaper]{article}

\usepackage[algorithms]{wacv}      


\definecolor{wacvblue}{rgb}{0.21,0.49,0.74}
\usepackage[pagebackref,breaklinks,colorlinks,allcolors=wacvblue]{hyperref}
\usepackage{graphicx}
\usepackage{amsmath}
\usepackage{amssymb}
\usepackage{booktabs}
\usepackage{pifont}
\usepackage{xcolor}
\usepackage[table]{xcolor}
\newcommand{\cmark}{\ding{51}}
\newcommand{\xmark}{\ding{55}}
\usepackage{multirow}
\usepackage{booktabs}
\usepackage{comment}

\usepackage{float}

\def\wacvPaperID{37} 
\def\confName{WACV}
\def\confYear{2026}

\title{Illuminating Darkness: Learning to Enhance Low-light Images In-the-Wild}

\author{
    S. M. A. Sharif$^{1}$ \quad
    Abdur Rehman$^{1}$ \quad
    Zain Ul Abidin$^{2}$ \\
    Fayaz Ali Dharejo$^{3}$ \quad
    Radu Timofte$^{3}$\thanks{Radu Timofte and Rizwan Ali Naqvi are the corresponding authors.} \quad
    Rizwan Ali Naqvi$^{2}$ \\
    $^1$Opt-AI Inc. \quad
    $^2$Sejong University \quad
    $^3$University of Würzburg \\ 
    \texttt{\href{https://github.com/sharif-apu/LSD-TFFormer}{Code and dataset: github.com/sharif-apu/LSD-TFFormer}} \vspace{-0.5cm}
}

\begin{document}



\twocolumn[{%
  \maketitle
  \begin{center}
    \includegraphics[width=\textwidth]{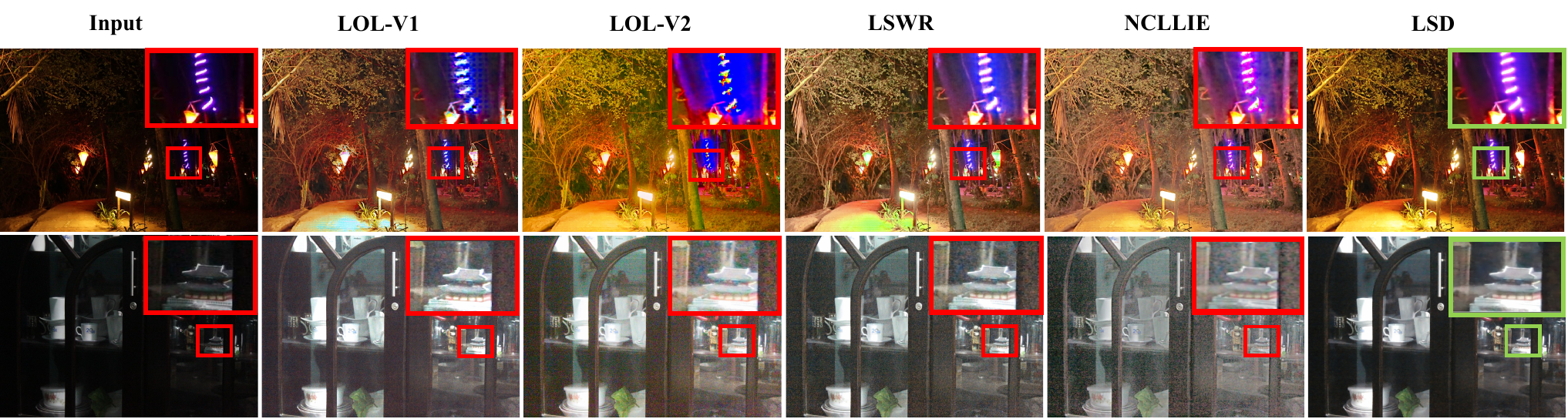}
    \captionof{figure}
    {Qualitative comparison of real-world low-light enhancement using our proposed TFFormer trained on different datasets. From left to right: Input image, TFFormer trained on LOL-V1~\cite{wei2018deep}, LOL-V2~\cite{yang2021sparse}, LSRW~\cite{hai2023r2rnet}, NCLLIE~\cite{liu2024ntire}, and the proposed LSD dataset. While existing datasets lead to artifacts or noise amplification, LSD enables robust enhancement with preserved structure and color fidelity, demonstrating its superior diversity and realism.}
    \label{fig:intro}
  \end{center}
  \vspace{1ex}
}]

\input{sec/0_abstract}    
\vspace{-0.2cm}

\input{sec/1_intro}
\input{sec/2_lsd}
\input{sec/3_tfformer}

\input{sec/4_experiments}
\input{sec/5_conclusion}
\input{sec/acknowledgement}
\appendix
\section*{Supplementary Material}
\input{sec/sup}

{
    \small
    \bibliographystyle{ieeenat_fullname}

}

\end{document}

%% file: sec/0_abstract.tex
\begin{abstract}
Single-shot low-light image enhancement (SLLIE) remains challenging due to the limited availability of diverse, real-world paired datasets. To bridge this gap, we introduce the Low-Light Smartphone Dataset (LSD), a large-scale, high-resolution (4K+) dataset collected in the wild across a wide range of challenging lighting conditions (0.1–200 lux). LSD contains 6,425 precisely aligned low and normal-light image pairs, selected from over 8,000 dynamic indoor and outdoor scenes through multi-frame acquisition and expert evaluation. To evaluate generalization and aesthetic quality, we collect 2,117 unpaired low-light images from previously unseen devices. To fully exploit LSD, we propose TFFormer, a hybrid model that encodes luminance and chrominance (LC) separately to reduce color-structure entanglement. We further propose a cross-attention-driven joint decoder for context-aware fusion of LC representations, along with LC refinement and LC-guided supervision to significantly enhance perceptual fidelity and structural consistency. TFFormer achieves state-of-the-art results on LSD (+2.45 dB PSNR) and substantially improves downstream vision tasks, such as low-light object detection (+6.80 mAP on ExDark). 

\end{abstract}

%% file: sec/1_intro.tex
\section{Introduction}
\label{sec:intro}

Learning-based SLLIE has gained significant momentum in recent times. It can improve the aesthetic appearance and significantly improve the performance of widely used vision tasks \cite{zhang2021beyond, zhang2021learning, sharif2023darkdeblur}. Unfortunately, models trained on existing SLLIE datasets, such as low-light (LOL) -V1 \cite{wei2018deep}, LOL-V2 \cite{yang2021sparse}, LSRW \cite{hai2023r2rnet}, and NTIRE challenge low-light enhancement (NCLLIE) \cite{liu2024ntire}, struggle to generalize to complex real-world environments. As a result, deep models often produce over-brightened areas with color shifts (Fig.~\ref{fig:intro}, top) and amplified noise (Fig.~\ref{fig:intro}, bottom).

\begin{figure}[!htb]
 \centering
 \begin{subfigure}[t]{0.495\linewidth}
 \centering
 \includegraphics[width=4.3cm]{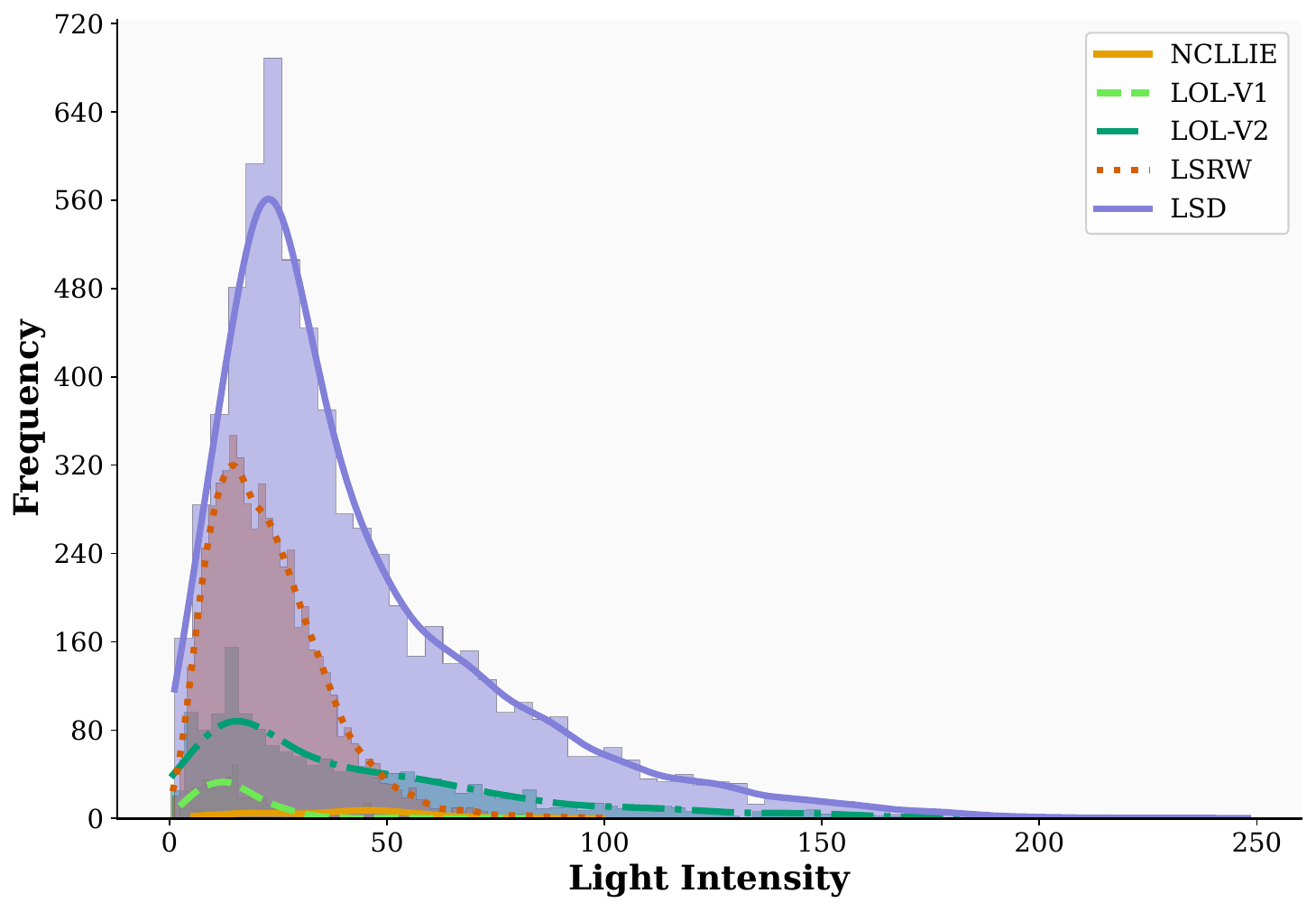}
 \caption{}
 \label{fig:noiseDataset}
 \end{subfigure}
 \begin{subfigure}[t]{0.495\linewidth}
 \centering
 \includegraphics[width=4.3cm]{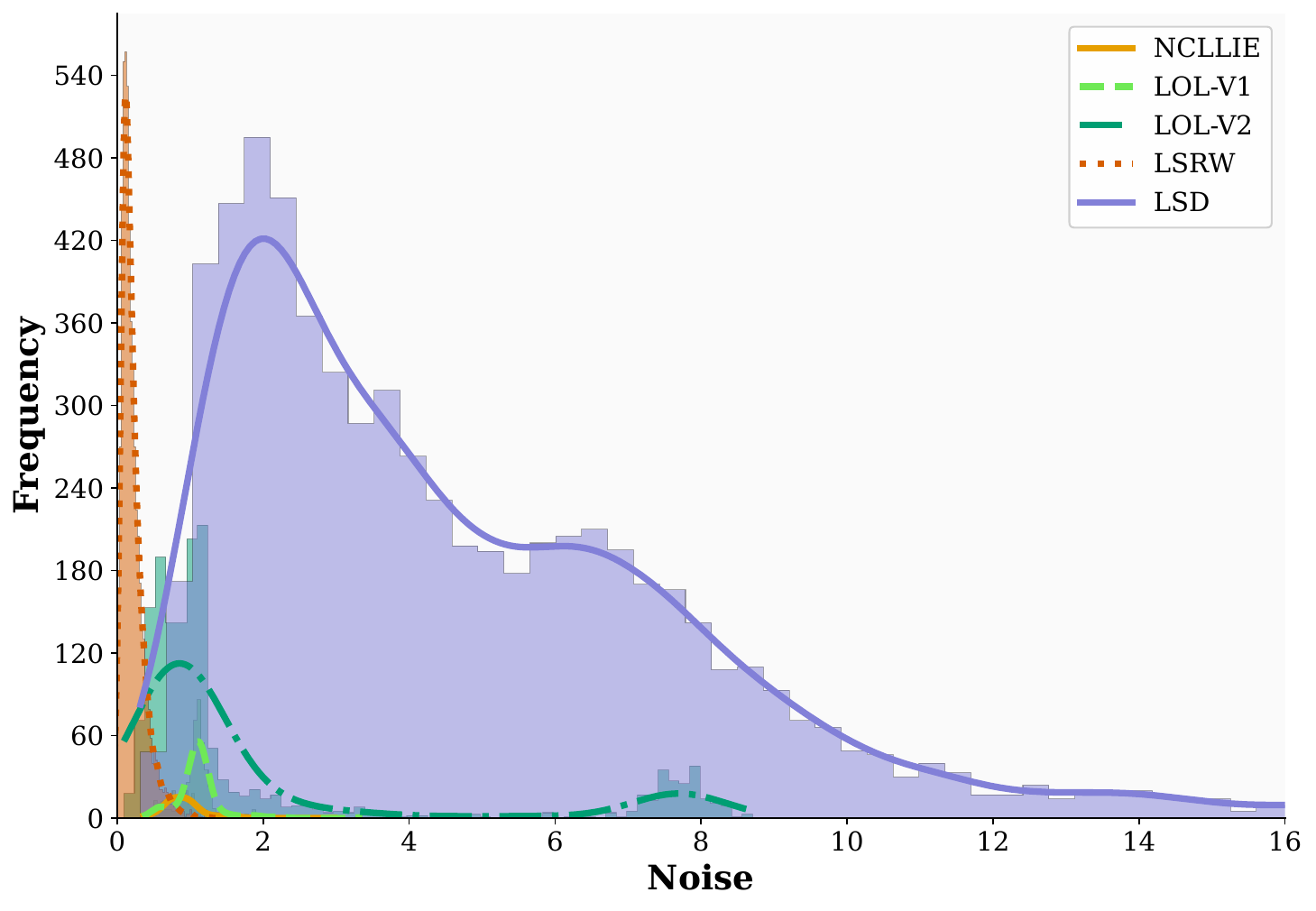}
 \caption{}
 \label{fig:intensityDataset}
 \end{subfigure}
 
 \caption{Noise and intensity distribution comparison between LSD and prior SLLIE datasets. (a) light intensity variation. (b) noise variation.}
 \label{fig:lsdvsslliedataset}
\end{figure}

The primary limitation of SLLIE stems from the characteristics of the training datasets. Existing datasets offer a limited number of training and evaluation samples captured in controlled environments \cite{xiong2020unsupervised}. These datasets often rely on synthetic low-light images created with low ISO settings (e.g., ISO 50/100 in LSRW \cite{hai2023r2rnet}) and artificially dimmed bright-light scenes. Therefore, they fail to represent critical real-world phenomena: sensor-induced noise, spatially varying light intensity, and color distortions that usually arise from the high-ISO amplification and short exposure times utilized to mitigate motion blurs in stochastic lighting conditions\cite{sharif2023darkdeblur, zhou2021delieve}. The constrained noise and intensity distributions of these datasets are illustrated in Fig.~\ref{fig:lsdvsslliedataset}, highlighting their significantly limited dynamic range.

To address these limitations, we introduce LSD, a large-scale real-world dataset comprising 6,425 high-resolution, focus-aligned image pairs captured using 15 different smartphone camera setups. The data spans diverse indoor (2,893) and outdoor (3,532) environments under uncontrolled lighting conditions ranging from 0.1 to 200 lux. Of these, 6,025 scenes are used to extract 179,694 training patches, while 400 complete scenes are reserved for evaluation. To the best of our knowledge, LSD is the largest publicly available real-world SLLIE dataset in terms of full-scene, high-resolution image pairs (see Table~\ref{tab:lsdvsld} for details). Notably, LSD offers significantly greater scene and sensor diversity compared to the previously largest SLLIE dataset, LSRW~\cite{hai2023r2rnet}, with 5,600 cropped patch pairs. To evaluate generalization beyond paired training, we introduce LSD-U, a set of 2,117 unpaired low-light images from unseen devices with diverse imaging characteristics. Models trained on LSD demonstrate strong generalization and substantially improve downstream tasks, achieving a +6.80 mAP gain in object detection on the ExDark dataset~\cite{loh2019getting}.

\input{tables/lsdvslliedataset}

While LSD addresses dataset-related limitations, existing SLLIE models also struggle under real-world degradations. RGB-based models often \cite{wang2019underexposed, xu2022snr,zamir2020learning, wang2022uformer, cui2205you, hou2023global,lore2017llnet, tao2017llcnn, shen2017msr, liu2023lae, lu2020deepselfie, li2023ldnet, xu2023low, fu2023learning, fan2023lacn, jin2023dnf} entangle semantic structures with signal-dependent noise, leading to color instability and texture loss. Retinex-inspired methods \cite{wei2018deep, zhang2021beyond, wang2019underexposed, cai2023retinexformer, yi2023diff, zhang2019kindling, Wang_2019_CVPR, singh2024illumination, zhang2019dual, zhao2021retinexdip} decouple illumination and reflectance but often misestimate illumination in complex scenes, resulting in halo artifacts. Recent non-RGB methods (e.g., HVI \cite{yan2025hvi}, LYT \cite{brateanu2025lyt}) attempt to separate luminance and chrominance, but they suffer from hue discontinuities and chroma noise due to limited cross-domain interaction. Please see Sec.~\ref{sec:comparison} for more detailed comparisons.

We propose TFFormer, a tuning-fork-shaped Transformer-CNN architecture that addresses key limitations in existing SLLIE methods. Unlike RGB-based~\cite{wang2022uformer, cui2205you, hou2023global} and Retinex-inspired~\cite{cai2023retinexformer, yi2023diff} models, TFFormer leverages structural-chromatic information by encoding luminance and chrominance separately, enabling domain-specific feature learning for structure-preserving denoising and color-consistent enhancement. To bridge the inter-domain correlation gaps observed in recent fixed color-space methods~\cite{yan2025hvi}, we introduce a Luminance–Chrominance Cross-Attention Block (LCCAB) and a joint decoder that enables adaptive fusion exclusively at the decoding stage. This late-stage interaction leverages spatially adaptive fusion, preserving domain-specific features while ensuring a structurally coherent reconstruction. Furthermore, we introduce the LC Guided Refinement Block (LCGRB) and LC Consistency Guidance loss (LCCG) to improve global structural fidelity and perceptual alignment. TFFormer outperforms the prior SOTA by +2.45 dB PSNR on LSD and generalizes well across diverse real-world scenes.

{\textbf{Contributions.}} Our main contributions are:
\begin{enumerate}
 \item We present LSD, the largest real-world SLLIE dataset, featuring over 6,400 paired samples and a 2,100-image unpaired benchmark for generalization testing.
 \item We propose TFFormer, a novel LC-decoupled hybrid Transformer-CNN architecture featuring LCCAB, LCGRB, and LCCG for perceptually consistent enhancement.
 \item We validate our method extensively with user studies, downstream vision task improvement, and generalization across unseen scenes.
\end{enumerate}

%% file: tables/lsdvslliedataset.tex
\begin{table}[!htb]
\centering
\rowcolors{1}{gray!15}{white}
\scalebox{0.49}{
\begin{tabular}{cccccccccc}
\toprule
Dataset & Sensors & ISO     & Image Size & Full Scene            & Outdoor & Indoor & Training & Testing & Total \\ \midrule
LOL-V1 \cite{wei2018deep} & 1       & -       & $400 \times 600$  & \textcolor{red}{\xmark }& 36      & 464    & 485      & 15      & 500   \\
LOL-V2 \cite{yang2021sparse} & 1       & -       & $400 \times 600$  & \textcolor{red}{\xmark }& 46      & 743    & 689      & 100     & 789   \\
LSRW   \cite{hai2023r2rnet} & 2       & 50,100  & $960\times720$    & \textcolor{red}{\xmark }& 846     & 4,804  & 5,600    & 50      & 5,650 \\
NCLLIE \cite{liu2024ntire} & 1       & -       & 4k+        & \textcolor{green}{\cmark }& 142     & 93     & 210      & 15      & 225   \\ \midrule
LSD     & 15      & 50-3200 & 4k+        & \textcolor{green}{\cmark }& 3,532   & 2,893  & \textbf{6,025}    & \textbf{400}     & \textbf{6,425} \\ \bottomrule
\end{tabular}}
 \caption{Details comparison between LSD and SLLIE datasets. The proposed LSD has the most diverse and most number of pair images collected with dynamic settings.}
  \label{tab:lsdvsld}
\end{table}

%% file: sec/2_lsd.tex

\begin{figure*}[!htb]
  \centering
  \begin{subfigure}[t]{0.6\linewidth}
      \includegraphics[width=\linewidth, height=5.8cm]{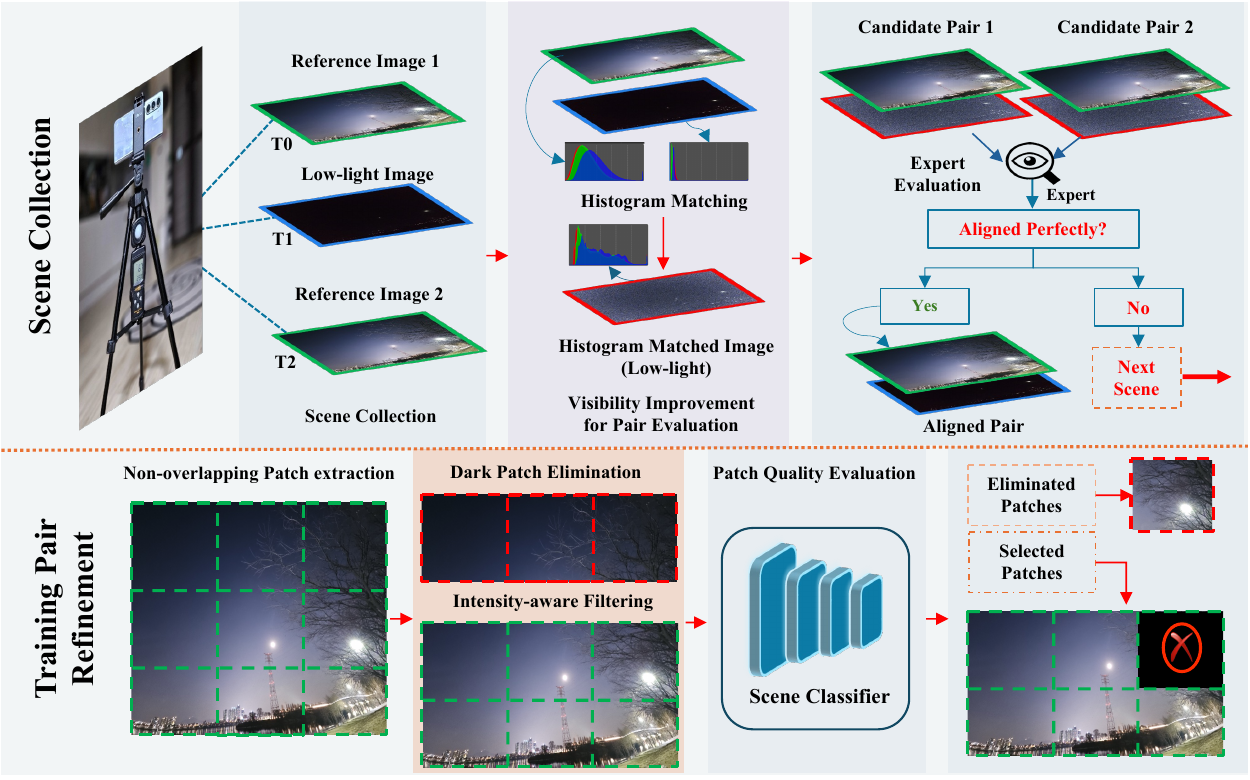}
      \caption{}
      \label{fig:lsdframework}
  \end{subfigure}
  \hfill
  \begin{subfigure}[t]{0.15\linewidth}
  \vspace{-5.82cm}
      \includegraphics[width=\linewidth, height=2.875cm]{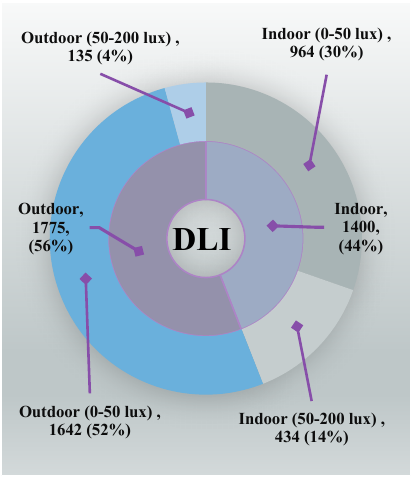}
      \includegraphics[width=\linewidth, height=2.875cm]{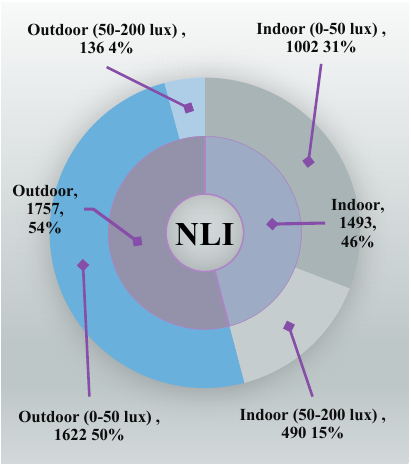}
      \caption{}
      \label{fig:lsddistribution}
  \end{subfigure}
  \hfill
  \begin{subfigure}[t]{0.24\linewidth}
      \includegraphics[width=\linewidth, height=5.8cm]{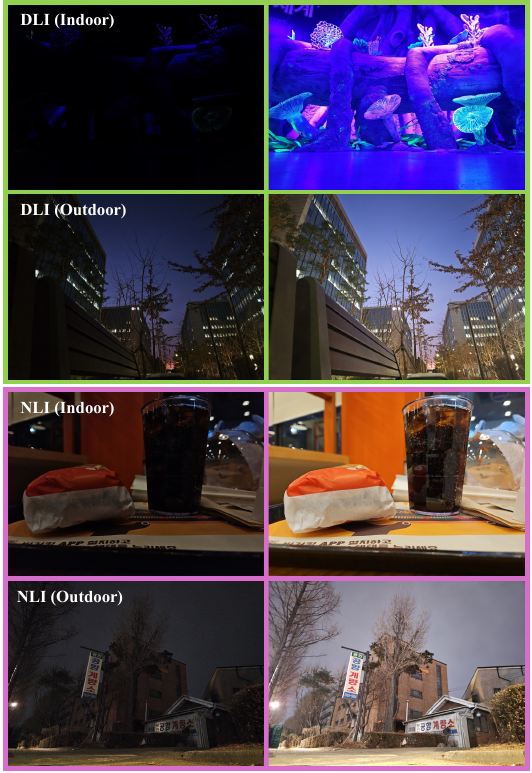}
      \caption{}
      \label{fig:lsdpairsample}
  \end{subfigure}
  \caption{Overview of the proposed LSD dataset:  (a) collection and refinement pipeline, (b)  Illumination distribution across indoor and outdoor scenes, (c) DLI and NLI sample pairs.}
\vspace{-0.2cm}
\end{figure*}



\section{Low-light Smartphone Dataset (LSD)}
\label{sec:lsd}

We present LSD, the largest SLLIE dataset collected in-the-wild. Fig. \ref{fig:lsdframework} illustrates the LSD collection pipeline.

\subsection{Scene acquisition}

\textbf{Collection Setup.} Capturing real-world aligned image pairs for SLLIE is inherently challenging due to motion artifacts and spatial disparities \cite{xiong2020unsupervised}. We collected more than 8,000 scenes using 15 camera setups. The scenes were captured in uncontrolled environments, spanning from sunset to sunrise. To fully exploit the proposed pipeline (Fig.~\ref{fig:lsdframework}), we developed an Android app with adjustable ISO, shutter speed, and denoising, leaving other settings on auto. To ensure spatial alignment, we adopted a three-frame capture strategy where two high-quality references bracketed each low-light image. To mitigate the effects of transient motion, we waited for stable moments, such as pauses between moving clouds or vehicles, before initiating capture. Additionally, our multi-frame setup allowed us to verify and reject scenes with observable motion inconsistencies. We acquired two input variants: (1)  denoise low-light input \textbf{(DLI)}: applied denoising to low-light images processed by the smartphone ISP, and (2) noisy low-light input \textbf{(NLI)}: raw noisy captures for enhancing diversity.

\noindent{\textbf{Capture Calibration.}} We employed a dynamic calibration strategy to capture low-light inputs and their corresponding reference images. For reference images, settings were calibrated to a low ISO range (i.e., 50–400) to minimize sensor noise. To achieve bright, naturally lit reference images, we used long exposure times, ranging from 1,000 to 15,000 ms, depending on the scene conditions. For low-light inputs, we utilized two distinct capture settings:
 \textbf{(1) ISP-guided}: ISP-driven settings: We took a dummy image in the stock camera app using auto mode before capturing each low-light scene. We then analyzed the sensitivity and shutter speed used in the dummy capture and applied these settings to the low-light input image to replicate camera characteristics. \textbf{(2) Random-sensitivity settings}: ISO and shutter speed configurations are hardware-dependent\cite{li2021low}. However, we aimed to generalize SLLIE across various scenes and devices. Thus, in approximately 60\% of scenes, we randomly adjusted ISO and shutter speed to introduce noise diversity while maintaining a low-light scene appearance similar to the dummy capture.

\noindent\textbf{Align-pair Selection.} Our LSD collection pipeline captures aligned image pairs by fixing the camera's focal point in each scene. However, we observed that smartphone hardware occasionally shifts the focal point (10-35\% of cases), requiring expert reviewers to evaluate spatial alignment and discard scenes with minor disparities. For each scene, we generated two candidate pairs by matching the low-light input with adjacent reference frames. To improve the visibility of extreme low-light inputs for evaluating spatial alignment, we applied histogram matching~\cite{shapira2013multiple, shen2007image} to boost visibility for manual expert review. Our experts discarded approximately 1,500 scenes with global misalignment. Fig.\ref{fig:lsdpairsample} shows examples of aligned (selected) DLI and NLI pairs.

\noindent\textbf{LSD Composition.} Following expert evaluation, we selected 6,425 aligned and in-focus scenes from the collected data to construct the LSD dataset, comprising high-resolution image pairs ranging from $3820\times2160$ to $4080\times3060$. The final dataset includes 2,893 indoor and 3,532 outdoor scenes captured under a broad range of illumination conditions. We categorize lighting into extreme low-light (0.1–50 lux) and low-light (50–200 lux) regimes. A total of 6,025 pairs are used for training, while 400 complete scenes are reserved for benchmarking. These benchmark scenes include both DLI and NLI variants across diverse lighting conditions and scene types, as illustrated in Fig.~\ref{fig:lsddistribution}. Further dataset details are provided in the \textbf{supplementary} material.

\begin{figure}[!htb]
  \centering
  \begin{subfigure}[t]{0.49\linewidth}
      \centering
      \includegraphics[width=\textwidth]{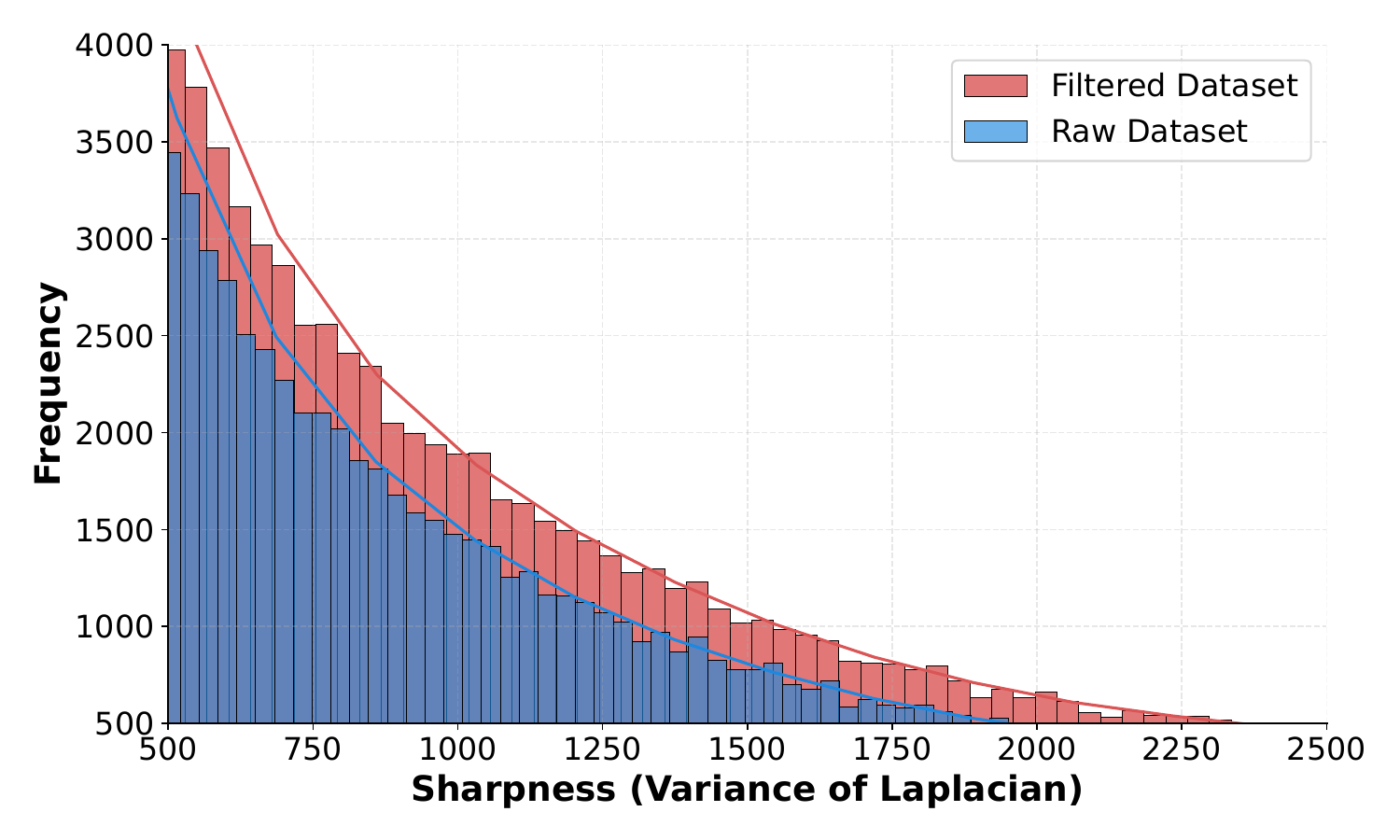}
      \caption{}
      \label{fig:sharpnessComp}
  \end{subfigure}
  \begin{subfigure}[t]{0.49\linewidth}
      \centering
      \includegraphics[width=\textwidth]{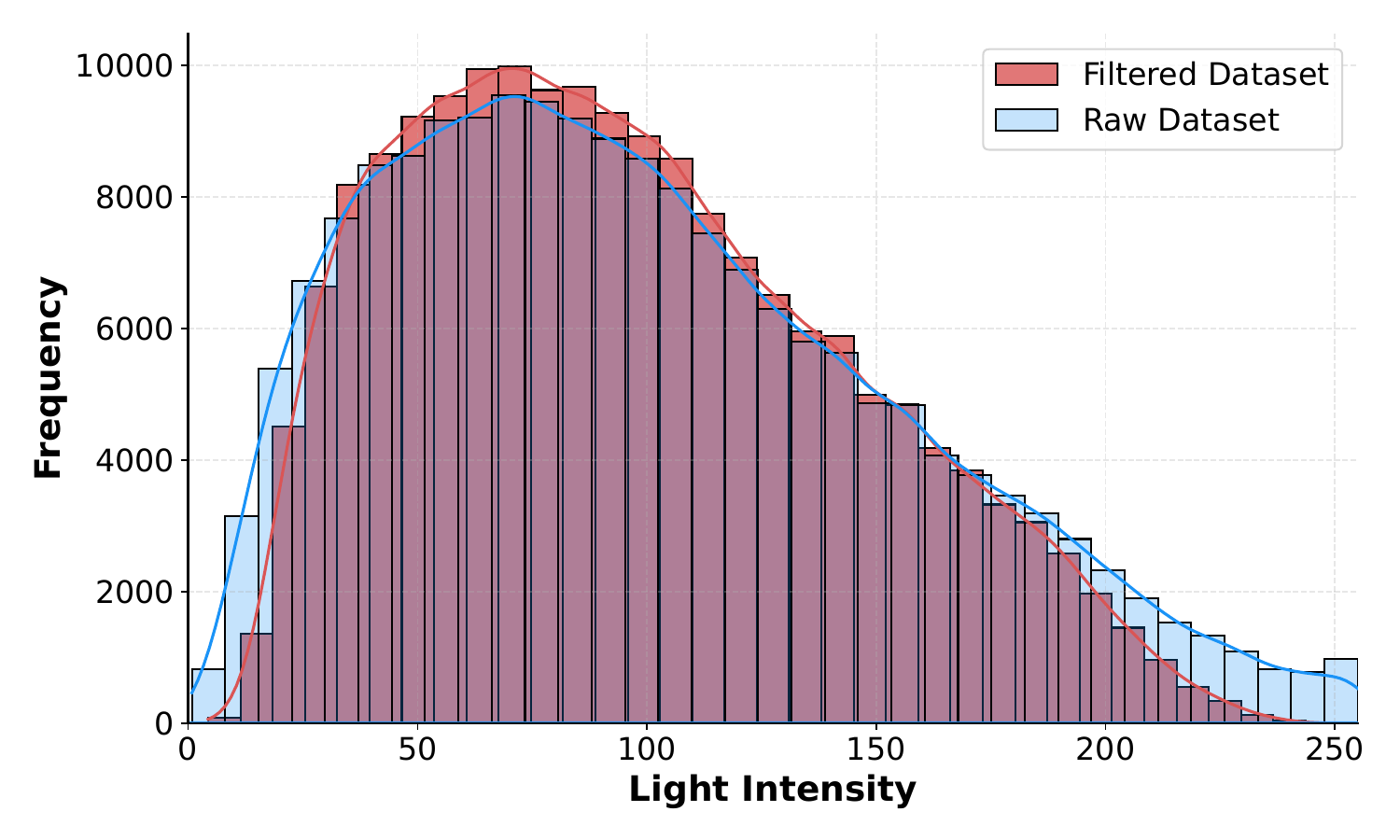}
      \caption{}
      \label{fig:intensityComp}
  \end{subfigure}

  \caption{Distribution comparison between filtered and raw training images. (a) sharpness improvement. (b) intensity filtering}
  \label{fig:lsdfilter}
\end{figure}

\subsection{Refining Training Pairs}

Despite controlled references, minor spatial degradations like defocus, local darkness, or overexposed highlights may persist~\cite{li2021low, liu2021benchmarking}. These issues can occasionally cause deep models to produce blurry regions and inconsistent exposure in extreme cases (see Sec. \ref{sec:abl}). Unlike existing works \cite{wei2018deep, hai2023r2rnet} that ignore such critical spatial evaluation, we introduce a reference-guided patch pair elimination method. Thus, we extracted $512\times512$ patches, yielding 208,853 pairs, then filter-out ~14\% using two filters:
\noindent\textbf{(1) Brightness threshold.} We discarded patches whose reference images had average intensity below $T = 10$, computed as
\begin{equation}
\scalebox{.95}{
    $I_{\text{avg}} = \frac{1}{N} \sum_{i=1}^{N} I_i$}
\end{equation}
This threshold filters out extremely dark reference patches that appear nearly black on standard displays and typically lack meaningful structure due to underexposure or ISP-induced smoothing. Removing these patches ensures the model learns from visually informative and reliable supervision.

\noindent\textbf{(2) Confidence filter.} We developed a VGG-based \cite{liu2015very} scene classifier to evaluate patch quality evaluation. The confidence score $C$ for each reference image, computed as $C = f(\mathbf{I_G})$ (where $\mathbf{I_G}$ represents the reference image), was used to filter out pairs with $C < 0.90$. Since reference images guide deep models in generating plausible outputs, we prioritized their quality during the filtering process. Fig. \ref{fig:lsdfilter} compares the distributions of the raw and filtered patches, highlighting the improvements in intensity and sharpness in the refined training set.





\subsection{LSD-U: Unpair Benchmarking Dataset}

SLLIE is fundamentally an aesthetic task, requiring methods to generalize across diverse real-world scenarios. Accordingly, evaluation should be performed on datasets with broad scene variety and data distributions, rather than those limited by specific biases. Thus, we introduce LSD-U, including 2,117 unique unpaired test cases. Our benchmark features data from DSLR cameras, smartphones, low-light video frames, and social media images sourced from various users with different photographic styles and setups compared to the proposed LSD. The LSD-U distribution introduces varied biases, enabling SLLIE methods to be tested across a wide range of real-world conditions, including enhancing trillions of social media images.

%% file: sec/3_tfformer.tex
\section{Learning with TFFormer}
\label{sec:TFFormer}

TFFormer learns to map $\mathrm{T}\!:\mathbf{I}_{LL}\!\rightarrow\!(\mathbf{I}_{rec},\mathbf{I}_{ref})$, where $\mathbf{I}_{LL}$ is the input low-light image, $\mathbf{I}_{rec}$ an intermediate reconstruction, and $\mathbf{I}_{ref}$ the final refined output. The pipeline includes: (1) LC mapping for LC decomposition, (2) LC encoders with attention-based encoding, (3) cross-attention-based LC feature fusion, (4) a joint decoder generating $\mathbf{I}_{rec}$, and (5) an LC Guided Reconstruction Block (LCGRB) refining $\mathbf{I}_{rec}$ into $\mathbf{I}_{ref}$. This modular design facilitates structured fusion and progressive refinement. Figure~\ref{fig:TFFormer} illustrates the pipeline.

\subsection{Luminance-Chrominance Mapping}
\label{LCExtraction}
\begin{figure*}[!htb]
  \centering
  \includegraphics[width=\linewidth]{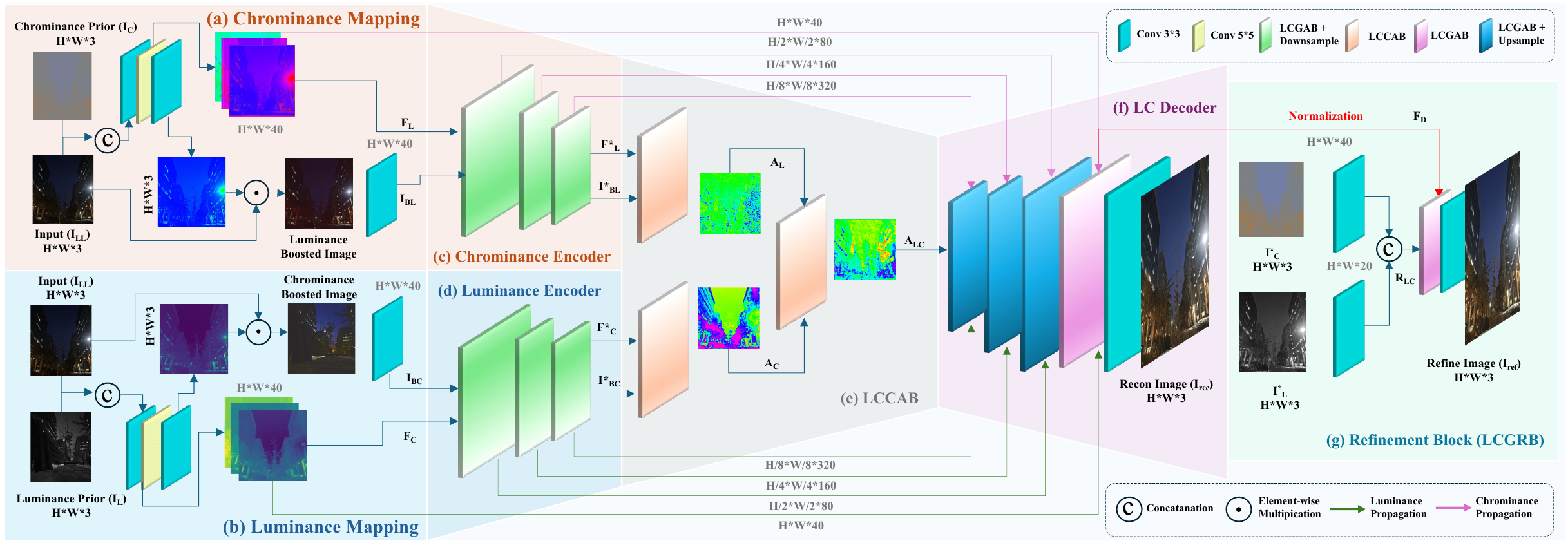}
  \caption{Overview of the proposed TFFormer architecture. (a), (b) LC mapping modules extract and boost luminance and chrominance features. (c), (d) Dedicated encoders process the respective LC features. (e) LC Cross-Attention Block (LCCAB) fuses luminance and chrominance representations. (f) A shared decoder reconstructs an intermediate output. (g) The LC Guided Refinement Block (LCGRB) further enhances the image using LC-aware attention.}

  \label{fig:TFFormer}
   \vspace{-0.3cm}
\end{figure*}

Given a low-light RGB image $\mathbf{I}_{LL}\in\mathbb{R}^{H\times W\times3}$, we separate it into a luminance map $\mathbf{I}_L$ and chrominance prior $\mathbf{I}_C$. The luminance is a weighted sum of the color channels following \cite{zhu2015shadow,lin2007face}, and the chrominance is obtained by subtracting: $\mathbf{I}_C = \mathbf{I}_{LL} -\mathbf{I}_L$. Injecting raw $\mathbf{I}_L$ preserves structural details and curbs over-brightening, while learning on $\mathbf{I}_C$ safeguards color fidelity and prevents desaturation. This separation enables independent noise suppression in structure-sensitive (luminance) and color-sensitive (chrominance) domains. Each component passes through an identical mapping block, $\mathrm{LC}{\text{map}}(\cdot)$, built from three \textit{conv}–BN–GELU layers (Fig.~\ref{fig:TFFormer} (a,b)). The module outputs a feature tensor and "boosted'' image:
\begin{equation}
\begin{aligned}
(\mathbf{F}_L,\mathbf{I}_{B_L}) &= \mathrm{LC}_{\text{map}}(\mathbf{I}_{LL},\mathbf{I}_L)\\
(\mathbf{F}_C,\mathbf{I}_{B_C}) &= \mathrm{LC}_{\text{map}}(\mathbf{I}_{LL},\mathbf{I}_C)
\end{aligned}
\end{equation}
\noindent
where $\mathbf{F}_L$, $\mathbf{F}_C$, $\mathbf{I}_{B_L}$, $\mathbf{I}_{B_C}$ $\in$ $\mathbb{R}^{H \times W \times 40}$. Retinex-based SLLIE models such as Retinexformer \cite{cai2023retinexformer} compute boosted images directly from illumination features, biasing toward luminance and yielding colour over-enhancement. Our decoupled design retains chrominance information, leading to balanced enhancement with fewer colour artefacts.

\subsection{LC Encoder-Decoder}
\label{sec:LCEncoderDecoder}
TFFormer adopts an LC Encoder-Decoder architecture, where luminance and chrominance features are encoded separately using modality-specific encoders and fused during decoding (Fig.~\ref{fig:TFFormer}(c,d)). The luminance and chrominance encoders, $\mathcal{E}_L(\cdot)$ and $\mathcal{E}_C(\cdot)$, comprise multiple LC Guided Attention Blocks (LCGABs), which extract structure and color-sensitive features guided by semantic cues. Given boosted inputs $(I_B, F)\in\mathbb{R}^{HW\times C}$ from the LC mapping module, each LCGAB splits $I_B$ into $k$ attention heads, $I_B^{i}\in\mathbb{R}^{HW\times d_k}$, where $d_k=C/k$. Each head computes query, key, and value matrices:
\begin{equation}
Q_i = I_B^{i} W_i^Q,\quad K_i = I_B^{i} W_i^K,\quad V_i = I_B^{i} W_i^V,
\end{equation}
where $W_i^Q$, $W_i^K$, $W_i^V \in \mathbb{R}^{d_k \times d_k}$ are head-specific parameters. The scaled dot-product attention is computed using a learnable scalar temperature $\alpha_i$~\cite{cai2023retinexformer}:
\begin{equation}
\mathrm{Attn}(I_B, F) = \operatorname{Softmax}\left(\tfrac{Q_i K_i^\top}{\alpha_i}\right)V_i \odot F.
\label{eq:attention}
\end{equation}
Here, $\odot$ denotes element-wise multiplication, where the guidance LC feature $F$ modulates the attended output. LC encoders produce enhanced feature sets:
\begin{equation}
\begin{aligned}
E_L &= \mathcal{E}_L(I_{B_L}, F_L) = (I_{B_L}, F_L), \\
E_C &= \mathcal{E}_C(I_{B_C}, F_C) = (I_{B_C}, F_C),
\end{aligned}
\label{eq:encoder}
\end{equation}
where $E_L, E_C \in \mathbb{R}^{\tfrac{H}{8} \times \tfrac{W}{8} \times 320}$ represent the refined luminance and chrominance features. A cross-attention module (Sec.~\ref{LCCAB}) merges $E_L$ and $E_C$ into a fused representation $A_{LC}$. The shared decoder upsamples $A_{LC}$ through a symmetric stack of LCGABs, integrating skip connections from $E_L$ and $E_C$ to preserve fine details. The decoder outputs the intermediate enhanced RGB image $I_{\text{rec}} \in \mathbb{R}^{H \times W \times 3}$ with adaptively fused luminance and chrominance.
\begin{equation}
I_{\text{rec}} = \mathcal{D}(A_{LC}, E_L, E_C),
\label{eq:decoder}
\end{equation}
where $\mathcal{D}(\cdot)$ denotes the joint decoder that performs attention-guided upsampling and fusion using the encoder outputs.

\subsection{LC Cross Attention Block (LCCAB)}
\label{LCCAB}

Separate luminance and chrominance encoding helps preserve structural details and enables domain-specific denoising. However, this can hinder cross-domain interaction, leading to spatial and chromatic inconsistencies. To address this, we introduce the LCCAB, which adaptively fuses the LC features for perceptual coherence. As shown in Fig.~\ref{fig:TFFormer} (e), LCCAB applies multi-head self-attention to luminance and chrominance tokens independently, producing $\mathbf{A_L} = \text{Attn}(\mathbf{E_L})$ and $\mathbf{A_C} = \text{Attn}(\mathbf{E_C})$. Cross-attention is then performed with $\mathbf{A_L}$ as queries and $\mathbf{A_C}$ as keys and values, resulting in the fused representation $\mathbf{A_{LC}} = \text{CA}(\mathbf{A_L}, \mathbf{A_C})$. This enhances inter-domain dependencies, improving illumination and chroma alignment in the reconstructed image $\mathbf{I_{rec}}$.

\subsection{LC Guided Reconstruction Block (LCGRB)} 


The LCGRB refines the intermediate image $\mathbf{I_{\mathbf{L_{Rec}}}} \to \mathbf{I_{Ref}}$ by extracting reconstructed luminance ($\mathbf{I^\ast_L}$) and chrominance ($\mathbf{I^\ast_C}$) components (Section \ref{LCExtraction}), mapped to $\mathbf{I^\ast_L}, \mathbf{I^\ast_C} \in \mathbb{R}^{H \times W \times 20}$ using $3 \times 3$ convolution refinement. They are concatenated to form $\mathbf{R_{LC}} \in \mathbb{R}^{H \times W \times 40}$. The LCGAB processes $\mathbf{R_{LC}}$ to produce the refined output $\mathbf{I_{ref}}$ (Fig. \ref{fig:TFFormer}(e)). This enables global reconstruction refinement by leveraging LC properties without additional training phases. Ablation studies in Section~\ref{sec:abl} validate the effectiveness of all proposed blocks.

\subsection{LC Consistency Guidance (LCCG)}
TFFormer leverages LC characteristics by employing a per-pixel \(\ell_1\) distance between predicted and ground-truth LC components to address over-smoothing~\citep{sharif2023darkdeblur}. Unlike traditional feature-level consistency losses, LCCG enables separate handling of structure and color information. This loss is applied to both the intermediate reconstruction \(I_{\text{rec}}\) and the final refinement \(I_{\text{ref}}\), ensuring consistency between stages. By enhancing color fidelity, sharpening structural details, and smoothing gradients, LCCG facilitates stable training and progressive refinement.
\begin{equation}
\scalebox{0.88}{$
\mathcal{D}_{\text{LC}} \;=\;
\frac{1}{N}\sum_{i=1}^{N}
\Bigl(
\lVert I^{\text{gt}}_{L,(i)} - I^{\text{rec}}_{L,(i)} \rVert_{1} \;+\;
\lVert I^{\text{gt}}_{C,(i)} - I^{\text{rec}}_{C,(i)} \rVert_{1}
\Bigr)$}
\label{eq:lc_distance}
\end{equation}
where $I^{\text{gt}}_{L}$ and $I^{\text{rec}}_{L}$ (respectively
$I^{\text{gt}}_{C}$ and $I^{\text{rec}}_{C}$) denote the luminance
(respectively chrominance) channels of the ground-truth and reconstructed
images, and $N$ is the number of pixels.
We reuse the same distance for the refinement output, giving the overall loss
\begin{equation}
\mathcal{L}_{\text{LC}}
=
\mathcal{D}_{\text{LC}}\!\bigl(I^{\text{gt}},\,I^{\text{rec}}\bigr)
+
\mathcal{D}_{\text{LC}}\!\bigl(I^{\text{gt}},\,I_{\text{ref}}\bigr)
\label{eq:lc_loss}
\end{equation}

%% file: sec/4_experiments.tex
\input{tables/quantcomp}
\section{Experiments}

\subsection{Comparison between SLLIE methods}
\label{sec:comparison}

\textbf{Experimental Setup.} First, we benchmarked SOTA methods on the proposed LSD. We included various SLLIE approaches: Retinex-based models (RetinexNet~\cite{wei2018deep}, Kind~\cite{zhang2019kindling}, Kind+\cite{zhang2021beyond}, DeepUPE\cite{wang2019underexposed}), transformer-based methods (Retinexformer~\cite{cai2023retinexformer}, IAT~\cite{cui2205you}, SNRNet~\cite{xu2022snr}), general image restoration models (Uformer~\cite{wang2022uformer}, MIRNet~\cite{zamir2020learning}), a diffusion-based method (GSAD~\cite{hou2023global}), a hybrid diffusion-Retinex model (Diff-Retinex~\cite{yi2023diff}), and non-RGB-based methods (LYT~\cite{brateanu2025lyt}, HVI~\cite{yan2025hvi}). To ensure fair comparison, we reproduced and verified each method’s reported results on their recommended datasets, then retrained all models on the proposed LSD. For comprehensive evaluation, all models were trained on both DLI and NLI scenes.


\noindent
\textbf{Quantitative Evaluation.} Tab.~\ref{tab:quantComp} shows the performance of SLLIE methods in different lighting and environments. Our proposed \textbf{TFFormer} achieves state-of-the-art results across all scenarios. Compared to the strong baseline \textbf{RetinexFormer}~\cite{cai2023retinexformer}, TFFormer yields average gains of \textbf{+2.04 dB} in PSNR, \textbf{+0.0738} in SSIM, and \textbf{-0.0357} in LPIPS. It also surpasses the recent \textbf{HVI}~\cite{yan2025hvi} method by \textbf{+2.26 dB} in PSNR, \textbf{+0.0859} in SSIM, and \textbf{-0.0452} in LPIPS, confirming its robustness across diverse real-world lighting conditions.

In addition to accuracy, \textbf{TFFormer} offers a favorable performance complexity trade-off. With just \textbf{5.87M} parameters and \textbf{34.6 GFLOPs}, it is significantly more efficient than recent methods like \textbf{Diff-Retinex}~\cite{yi2023diff} (\textbf{80.7M}, \textbf{492.5 GFLOPs}) over \textbf{13$\times$} larger and \textbf{14$\times$} more computational yet achieves markedly lower performance. TFFormer outperforms it by \textbf{+3.56 dB} in PSNR, \textbf{+0.2005} in SSIM, and \textbf{-0.1206} in LPIPS. Compared to the heaviest model, \textbf{MIRNet}~\cite{zamir2020learning} (\textbf{31.79M}, \textbf{816 GFLOPs}), TFFormer is \textbf{5.4$\times$} smaller and \textbf{24$\times$} faster, while achieving \textbf{+3.83 dB} higher PSNR, \textbf{+0.1088} higher SSIM, and \textbf{-0.0417} lower LPIPS. These results underline TFFormer’s efficiency without compromising quality

\begin{figure*}[!htb]
  \centering
  \includegraphics[width=\textwidth]{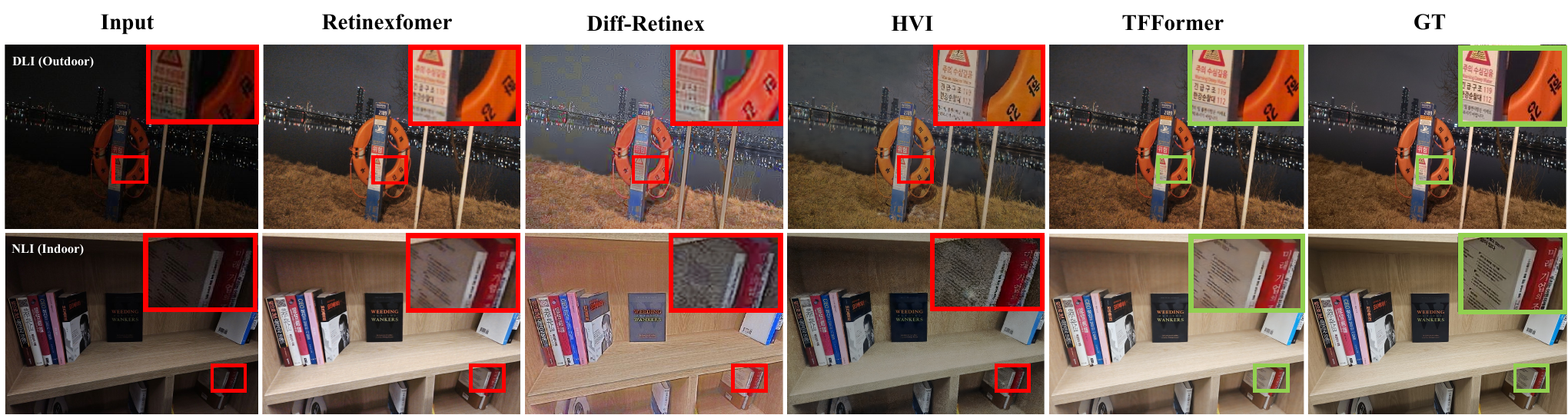}
  \caption{Qualitative comparison of existing SLLIE methods on the LSD dataset. The top scenes show performance on DLI, and the bottom example illustrates performance on DLI scenes. TFFormer produces cleaner, more plausible images, outperforming existing methods.}
  \label{fig:visComp}
\end{figure*}

\noindent
\textbf{Qualitative Evaluation.} Fig.~\ref{fig:visComp} visually compares SLLIE methods on the LSD. Retinex-based models like Diff-Retinex~\cite{yi2023diff} and Retinexformer~\cite{cai2023retinexformer} show color distortions. Diff-Retinex introduces yellow-green casts in NLI (Outdoor), while Retinexformer oversaturates warm tones in DLI (Outdoor). HVI~\cite{yan2025hvi} yields dim, desaturated outputs and fails to restore brightness and contrast, particularly in NLI (Indoor) environments. Structural fidelity also suffers: Diff-Retinex and HVI blur textures like wall gradients and bookshelf edges, and Retinexformer shows minor color bleeding that softens boundaries. In contrast, TFFormer recovers natural color, realistic brightness, and sharp structure, avoiding overenhancement in DLI and effectively denoising complex NLI scenes. The results confirm TFFormer's robustness and generalization across both noisy and clean low-light scenes.

\subsection{Cross-dataset Evaluation}

\subsubsection{Perceptual Comparison}

\begin{figure}[!htb]
  \centering
  \includegraphics[width=\linewidth, height=4cm]{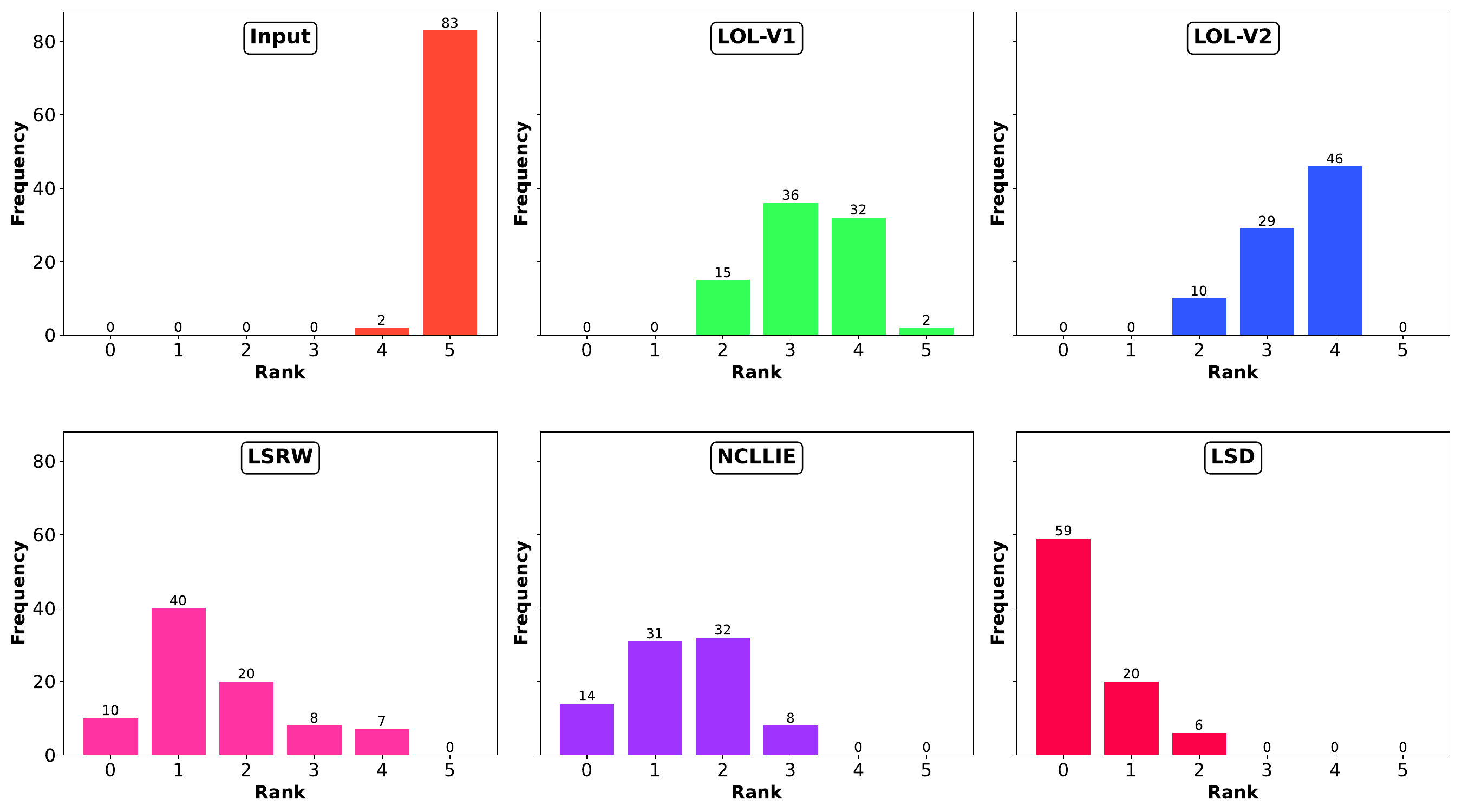}
  \caption{User rankings, summarized via the Bradley-Terry model, show a substantial margin preferred LSD outputs.}
  \label{fig:crossVis}
\end{figure}

The proposed LSD is designed to address diverse real-world low-light scenarios. For fair assessment of generalization and user preference, we performed cross-dataset evaluation using the independently collected LSD-U set. Notably, LSD-U features a distribution entirely distinct from LSD and all existing SLLIE datasets. We trained TFFormer separately on existing SLLIE datasets and LSD, then evaluated their acceptance among mass users through a perceptual user study. We recruited \textbf{15} participants, each presented with \textbf{5 randomly selected sets} from the 2,117-image LSD-U collection. Participants ranked the enhancement outputs based on visual preference. Rankings were aggregated using the Bradley-Terry model \cite{bradley1952rank}, revealing that TFFormer trained on LSD consistently outperformed counterparts trained on prior datasets (Fig.~\ref{fig:crossVis}).

\subsubsection{Vision Tasks}

Beyond perceptual quality, a robust SLLIE dataset should facilitate downstream vision tasks in low-light conditions. To evaluate this, we enhanced the ExDark validation set~\cite{loh2019getting}, which includes 1,473 images across 13 object categories, using LSD and existing SLLIE datasets. We then performed object detection with YOLOv3~\cite{farhadi2018yolov3}, pretrained on the COCO dataset~\cite{lin2014coco}. As shown in Tab.~\ref{tab:odcom}, LSD-enhanced images achieved the highest average detection performance at \textbf{38.80 mAP}, surpassing all other datasets. LSD also yielded the best or second-best results in most categories, notably improving detection for challenging classes such as Car, Cat, Chair, Cup, Dog, Motorbike, People, and Table. Compared to raw ExDark images, LSD-enhanced inputs provided a substantial \textbf{+6.80 mAP} gain, demonstrating the effectiveness in real-world low-light vision applications.

\input{tables/od_cross_eval}

\subsection{Generalization across devices}
\begin{figure}[!htb]

  \centering
  \includegraphics[width=\linewidth]{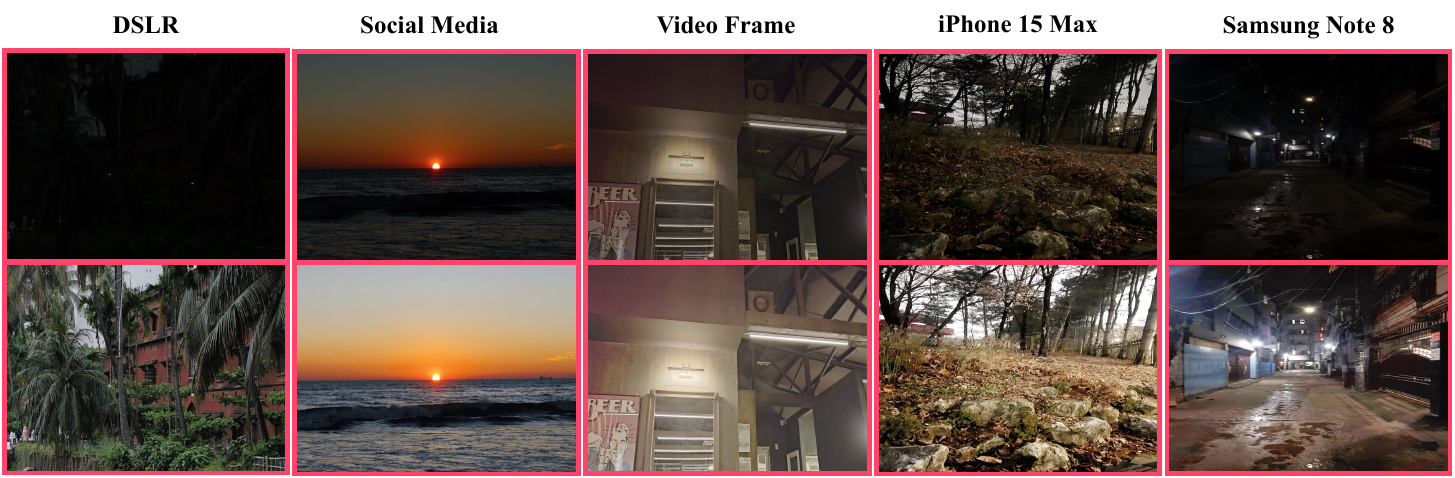}
  \caption{Performance of LSD-TFFormer on diverse sources, including old, heavily compressed social media images.}
  \label{fig:lsd-tfformer}

\end{figure}

Fig. \ref{fig:lsd-tfformer} shows the performance of LSD-TFFormer across various scenarios. Despite being trained on single-shot images, our model adapts well to video frames and can enhance photos taken on older devices, such as the eight-year-old Samsung Note 8. LSD-TFFormer can also effectively revive decade-old compressed images from social media images without visible artifacts. A separate user study confirmed the real-world effectiveness of the proposed method, with over \textbf{82\%} of participants preferring the enhanced images over their low-light counterparts. Please refer to the \textbf{supplementary} materials for more details.

\subsection{Ablation Study}
\label{sec:abl}
\begin{figure}[!htb]
  \centering
  \includegraphics[width=\linewidth]{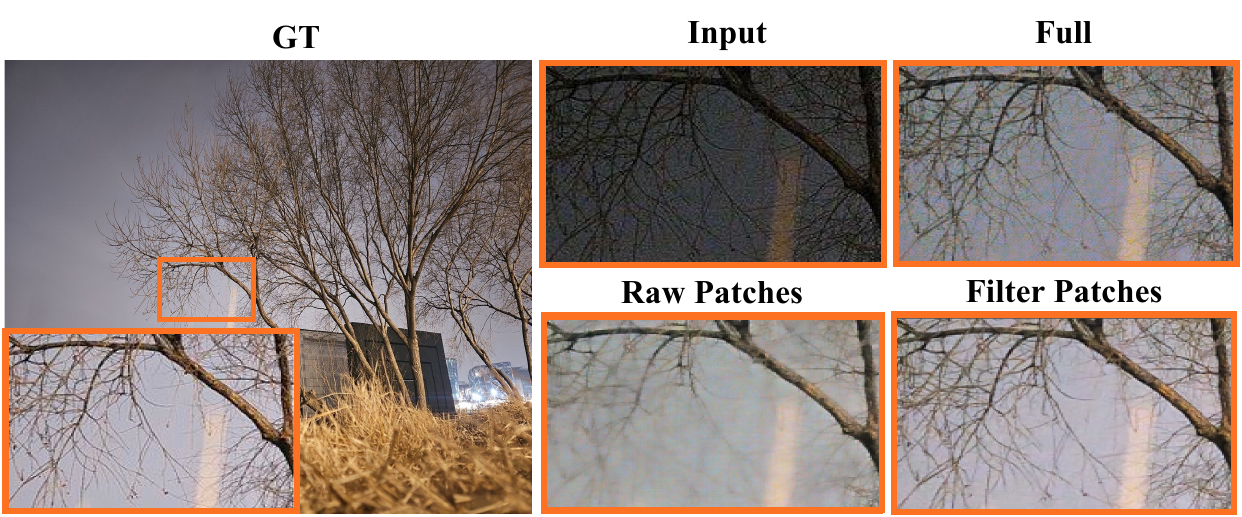}
  \caption{Ablation study on LSD, patch, and full-scene training}
  \label{fig:lsdAbl}

\end{figure}

\textbf{LSD.} Fig.~\ref{fig:lsdAbl} and Tab.~\ref{tab:abllsd} show TFFormer’s performance with three training strategies: raw patches, full-scene images, and our proposed filtered patches. Filtered patch training yields the best results, improving PSNR by \textbf{2.14 dB} on DLI and \textbf{1.62 dB} on average compared to full-scene training. By focusing on structurally rich and clean regions, it enables the model to learn localized noise patterns and preserve fine details, avoiding spatial blurs introduced by motion in GT images in extreme cases. In contrast, raw patch training underperforms due to exposure to noisy or misaligned content, leading to degraded output quality.

\input{tables/abl_lsd}

\begin{figure}[!htb]
  \centering
  \includegraphics[width=\linewidth]{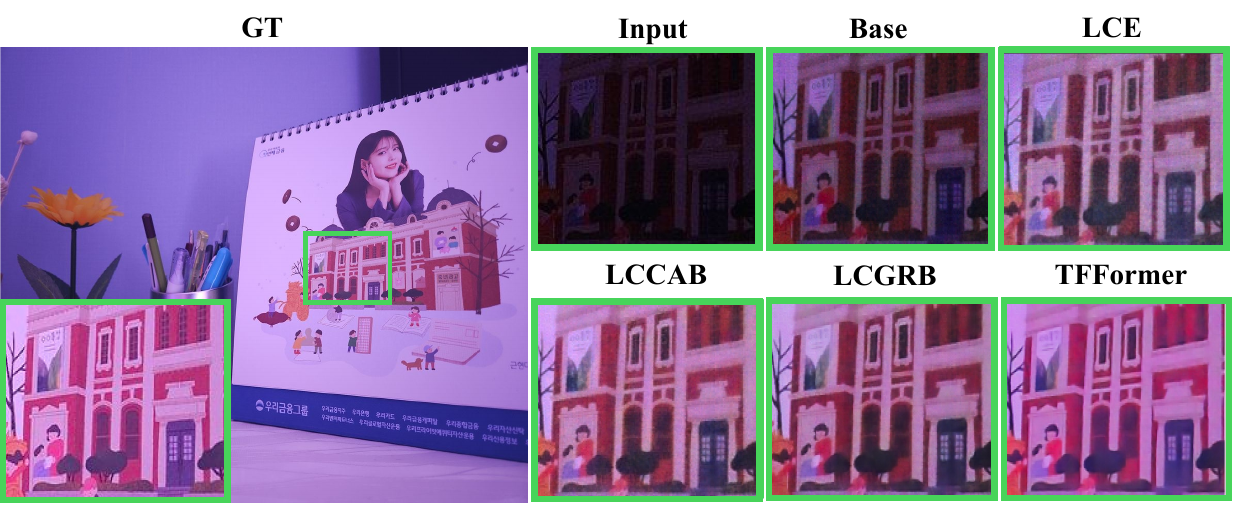}
  \caption{Ablation study on the novel component of the proposed TFFormer on LSD.}
  \label{fig:tffAbl}

\end{figure}

\textbf{TFFormer.} Tab.~\ref{tab:abltff} and Fig.~\ref{fig:tffAbl} present an ablation study evaluating each proposed component in TFFormer. Starting from a base model, we progressively add the decoupled LC Encoder (LCE), Cross-Attention  (LCCAB), Refinement Block (LCGRB), and Consistency Guidance (LCCG). LCE enhances denoising and color consistency by separating luminance and chrominance early in the image processing pipeline. Adding LCCAB enables adaptive inter-domain fusion with joint decoding, boosting global reconstruction. LCGRB enhances structural detail recovery, while LCCG improves perceptual alignment. The full model achieves a \textbf{+3.57 dB} PSNR gain and a \textbf{-0.0268} LPIPS reduction over the base, validating the effectiveness of our design for structure-aware and color-consistent enhancement in complex lighting conditions.

\input{tables/abl_tf}

\subsection{Discussion}
The proposed method validates a generic SLLIE framework and highlights limitations in existing datasets and models under diverse real-world conditions. Full implementation details, including sensor specs, app design, training strategies, and extended results, are provided in the \textbf{supplementary material}. While our model generalizes well, it struggles under extreme low-light (e.g., <1 lux), which we plan to address through joint denoising and enhancement in future work. We also aim to extend LSD to support downstream tasks like object detection and segmentation for practical applications.

%% file: tables/quantcomp.tex
\begin{table*}[!htb]
\centering
\rowcolors{1}{gray!15}{white}
\scalebox{.55}{

\begin{tabular}{ccccccccccccccccccc}
\toprule
                                                           &                         & \multicolumn{2}{c}{}                            & \multicolumn{6}{c}{Low-light}                                                                                                                                                               & \multicolumn{6}{c}{Extreme Low-light}                                                                                                                                                       & \multicolumn{3}{c}{LSD}                                                                      \\ \cmidrule(lr){3-4}\cmidrule(lr){5-10}\cmidrule(lr){11-16}\cmidrule(lr){17-19}
                                                           &                         & \multicolumn{2}{c}{\multirow{-1}{*}{Compexity}} & \multicolumn{3}{c}{Indoor}                                                                   & \multicolumn{3}{c}{Outdoor}                                                                  & \multicolumn{3}{c}{Indoor}                                                                   & \multicolumn{3}{c}{Outdoor}                                                                  & \multicolumn{3}{c}{Average}                                                                  \\ \cmidrule(lr){3-4}\cmidrule(lr){5-10}\cmidrule(lr){11-16}\cmidrule(lr){17-19}
\multirow{-3}{*}{Model}                                    & \multirow{-3}{*}{Scene} & Flops (G)              & Param. (M)             & PSNR $\uparrow$                        & SSIM $\uparrow$                          & LPIPS $\downarrow$                        & PSNR $\uparrow$                        & SSIM $\uparrow$                          & LPIPS $\downarrow$                        & PSNR $\uparrow$                        & SSIM $\uparrow$                          & LPIPS $\downarrow$                        & PSNR $\uparrow$                        & SSIM $\uparrow$                          & LPIPS $\downarrow$                        & PSNR $\uparrow$                        & SSIM $\uparrow$                          & LPIPS $\downarrow$                        \\ \midrule
Retinexnet \cite{wei2018deep}            &                         & 136.00                 & 0.84                   & 14.57                        & 0.6327                        & 0.2611                        & 13.70                        & 0.5489                        & 0.3150                        & 15.63                        & 0.5754                        & 0.3064                        & 14.34                        & 0.4871                        & 0.3875                        & 14.56                        & 0.5610                        & 0.3175                        \\
Kind \cite{zhang2019kindling}            &                         & 64.73                  & 8.01                   & 13.55                        & 0.6717                        & 0.2665                        & 13.47                        & 0.6287                        & 0.3174                        & 14.21                        & 0.6320                        & 0.3001                        & 13.25                        & 0.6047                        & 0.3521                        & 13.62                        & 0.6343                        & 0.3090                        \\
Kind+ \cite{zhang2021beyond}             &                         & 59.41                  & 8.27                   & 13.49                        & 0.6630                        & 0.2871                        & 13.17                        & 0.6138                        & 0.3411                        & 13.94                        & 0.6141                        & 0.3275                        & 13.13                        & 0.5899                        & 0.3733                        & 13.43                        & 0.6202                        & 0.3322                        \\
DeepUPE \cite{wang2019underexposed}      &                         & 0.04                   & 0.59                   & 15.73                        & 0.5874                        & 0.3008                        & 15.03                        & 0.5620                        & 0.2841                        & 14.99                        & 0.4612                        & 0.4104                        & 16.30                        & 0.4378                        & 0.4131                        & 15.51                        & 0.5121                        & 0.3521                        \\
SNRNet \cite{xu2022snr}                  &                         & 23.96                  & 39.12                  & 15.14                        & 0.6326                        & 0.3776                        & 14.82                        & 0.3510                        & 0.7161                        & 13.05                        & 0.5377                        & 0.4450                        & 15.43                        & 0.4646                        & 0.5720                        & 14.61                        & 0.4965                        & 0.5277                        \\
MIRNet \cite{zamir2020learning}          &                         & 816.00                 & 31.79                  & 16.15                        & 0.7206                        & 0.2284                        & 14.01                        & 0.5600                        & 0.3632                        & 15.51                        & 0.6360                        & 0.2700                        & 17.20                        & 0.5811                        & 0.3085                        & 15.72                        & 0.6244                        & 0.2925                        \\
Uformer \cite{wang2022uformer}           &                         & 12.00                  & 5.29                   & 13.04                        & 0.6578                        & 0.2672                        & 12.40                        & 0.5423                        & 0.3635                        & 11.93                        & 0.5575                        & 0.3092                        & 13.46                        & 0.5006                        & 0.3593                        & 12.71                        & 0.5646                        & 0.3248                        \\
IAT \cite{cui2205you}                    &                         & 1.44                   & 0.09                   & 13.34                        & 0.6557                        & 0.2819                        & 13.31                        & 0.4680                        & 0.4320                        & 14.82                        & 0.4978                        & 0.3944                        & 13.07                        & 0.5600                        & 0.3583                        & 13.64                        & 0.5454                        & 0.3667                        \\
Retinexformer \cite{cai2023retinexformer}&                         & 39.16                  & 3.38                   & 16.83                        & 0.7238                        & 0.2286                        & 15.76                        & 0.5900                        & 0.3551                        & 17.97                        & 0.6746                        & 0.2495                        & {\color[HTML]{3531FF} 17.73} & 0.5945                        & 0.3123                        & 17.07                        & 0.6457                        & 0.2864                        \\
GSAD \cite{hou2023global}                &                         & 67.02                  & 17.17                  & {\color[HTML]{3531FF} 18.31} & {\color[HTML]{3531FF} 0.7307} & {\color[HTML]{3531FF} 0.2195} & {\color{blue} 16.55} & {\color{blue} 0.5934} & {\color{blue} 0.2916} & {\color[HTML]{3531FF} 18.52} & {\color[HTML]{3531FF} 0.7042} & {\color[HTML]{3531FF} 0.2425} & 17.06                        & {\color[HTML]{3531FF} 0.6232} & {\color[HTML]{3531FF} 0.2993} & {\color[HTML]{3531FF} 17.61} & {\color[HTML]{3531FF} 0.6629} & {\color[HTML]{3531FF} 0.2632} \\
Diff-Retinex \cite{yi2023diff}           &                         & 492.47                 & 80.70                  & 14.85                        & 0.6235                        & 0.2795                        & 14.89                        & 0.5229                        & 0.3663                        & 16.16                        & 0.5993                        & 0.3302                        & 16.35                        & 0.5070                        & 0.4143                        & 15.56                        & 0.5632                        & 0.3476                        \\
LYT \cite{brateanu2025lyt}               &                         & 1.70                   & 0.04                   & 17.26                        & 0.7103                        & 0.2122                        & 16.41                        & 0.5533                        & 0.3776                        & 17.31                        & 0.6386                        & 0.2789                        & 18.41                        & 0.5725                        & 0.3363                        & 17.35                        & 0.6187                        & 0.3012                        \\
HVI \cite{yan2025hvi}                    &                         & 8.13                   & 1.97                   & 17.30                        & 0.7200                        & 0.2325                        & 16.68                        & 0.5690                        & 0.3006                        & 16.63                        & 0.6692                        & 0.2860                        & 17.63                        & 0.6009                        & 0.3325                        & 17.06                        & 0.6398                        & 0.2879                        \\ \midrule
TFFormer                                                   & \multirow{-14}{*}{DLL}  & 34.59                  & 5.87                   & {\color{red} 19.28} & {\color{red} 0.7766} & {\color{red} 0.2083} & {\color{red} 17.71} & {\color{red} 0.6576} & {\color{red} 0.3061} & {\color{red} 19.82} & {\color{red} 0.7363} & {\color{red} 0.2335} & {\color{red} 19.70} & {\color{red} 0.6660} & {\color{red} 0.2954} & {\color{red} 19.13} & {\color{red} 0.7091} & {\color{red} 0.2609} \\ \midrule
Retinexnet \cite{wei2018deep}            &                         & 136.00                 & 0.84                   & 14.11                        & 0.5456                        & 0.3465                        & 13.21                        & 0.5234                        & 0.3710                        & 14.44                        & 0.4252                        & 0.5545                        & 13.50                        & 0.4078                        & 0.5077                        & 13.81                        & 0.4755                        & 0.4449                        \\
Kind \cite{zhang2019kindling}            &                         & 64.73                  & 8.01                   & 13.49                        & 0.5330                        & 0.3495                        & 13.09                        & 0.5701                        & 0.3280                        & 13.43                        & 0.4331                        & 0.4819                        & 13.13                        & 0.5532                        & 0.3699                        & 13.28                        & 0.5223                        & 0.3823                        \\
Kind+ \cite{zhang2021beyond}             &                         & 59.41                  & 8.27                   & 13.22                        & 0.5283                        & 0.3662                        & 12.70                        & 0.5667                        & 0.3405                        & 13.62                        & 0.4214                        & 0.4949                        & 12.90                        & 0.4476                        & 0.4625                        & 13.11                        & 0.4910                        & 0.4160                        \\
DeepUPE \cite{wang2019underexposed}      &                         & 0.04                   & 0.59                   & 14.11                        & 0.3529                        & 0.5667                        & 13.42                        & 0.4061                        & 0.4229                        & 12.79                        & 0.2125                        & 0.7857                        & 14.01                        & 0.2799                        & 0.6351                        & 13.58                        & 0.3128                        & 0.6026                        \\
SNRNet \cite{xu2022snr}                  &                         & 23.96                  & 39.12                  & 16.66                        & 0.5793                        & 0.4208                        & 14.95                        & 0.3025                        & 0.7497                        & 14.04                        & 0.4882                        & 0.5207                        & 16.60                        & 0.4110                        & 0.6183                        & 15.56                        & 0.4453                        & 0.5774                        \\
MIRNet \cite{zamir2020learning}          &                         & 816.00                 & 31.79                  & 14.82                        & 0.6551                        & 0.2523                        & 13.64                        & 0.5297                        & 0.3274                        & 12.80                        & 0.5291                        & 0.3615                        & 16.74                        & 0.5524                        & {\color{blue} 0.3300} & 14.50                        & 0.5666                        & 0.3178                        \\
Uformer \cite{wang2022uformer}           &                         & 12.00                  & 5.29                   & 13.85                        & 0.6570                        & 0.2628                        & 13.07                        & 0.5654                        & 0.3315                        & 12.53                        & 0.5591                        & 0.3488                        & 14.13                        & 0.5325                        & 0.3609                        & 13.39                        & 0.5785                        & 0.3260                        \\
IAT \cite{cui2205you}                    &                         & 1.44                   & 0.09                   & 15.35                        & 0.6506                        & 0.2612                        & 13.92                        & 0.5248                        & 0.3434                        & 13.33                        & 0.5308                        & 0.3980                        & 15.88                        & 0.5062                        & 0.3671                        & 14.62                        & 0.5531                        & 0.3424                        \\
Retinexformer \cite{cai2023retinexformer}&                         & 39.16                  & 3.38                   & {\color[HTML]{3531FF} 17.16} & {\color[HTML]{3531FF} 0.6903} & {\color[HTML]{3531FF} 0.2493} & 15.14                        & 0.5908                        & 0.3297                        & 17.22                        & 0.6216                        & 0.3140                        & {\color{blue} 17.40} & 0.5585                        & 0.3549                        & 16.73                        & {\color{blue} 0.6153} & 0.3120                        \\
GSAD \cite{hou2023global}                &                         & 67.02                  & 17.17                  & 17.06                        & 0.6232                        & 0.2993                        & {\color{blue} 16.21} & {\color{blue} 0.5806} & {\color{blue} 0.2929} & {\color{blue} 17.89} & {\color{blue} 0.6601} & {\color{blue} 0.2907} & 16.45                        & {\color{blue} 0.5723} & 0.3589                        & {\color{blue} 16.90} & 0.6091                        & {\color[HTML]{3531FF} 0.3104} \\
Diff-Retinex \cite{yi2023diff}           &                         & 492.47                 & 80.70                  & 14.73                        & 0.4990                        & 0.3743                        & 14.87                        & 0.4745                        & 0.3342                        & 15.13                        & 0.3905                        & 0.5243                        & 16.05                        & 0.4131                        & 0.4494                        & 15.20                        & 0.4443                        & 0.4206                        \\
LYT \cite{brateanu2025lyt}               &                         & 1.70                   & 0.04                   & 16.33                        & 0.5886                        & 0.2926                        & 15.72                        & 0.5070                        & 0.3837                        & 16.27                        & 0.4915                        & 0.4289                        & 16.71                        & 0.4507                        & 0.4172                        & 16.26                        & 0.5095                        & 0.3806                        \\
HVI \cite{yan2025hvi}                    &                         & 8.13                   & 1.97                   & 16.43                        & 0.6593                        & 0.2816                        & 15.77                        & 0.5455                        & 0.3161                        & 16.55                        & 0.6384                        & 0.3360                        & 16.42                        & 0.5443                        & 0.3842                        & 16.29                        & 0.5969                        & 0.3295                        \\ \midrule
TFFormer                                                   & \multirow{-14}{*}{NLL}  & 34.59                  & 5.87                   & {\color{red} 18.99} & {\color{red} 0.7470} & {\color{red} 0.2231} & {\color{red} 17.28} & {\color{red} 0.6799} & {\color{red} 0.2823} & {\color{red} 18.94} & {\color{red} 0.7113} & {\color{red} 0.2571} & {\color{red} 19.75} & {\color{red} 0.6598} & {\color{red} 0.3018} & {\color{red} 18.74} & {\color{red} 0.6995} & {\color{red} 0.2661} \\ \bottomrule
\end{tabular}
}
\caption{Quantitative Benchmark on LSD Dataset. Our proposed TFFormer outperforms the existing methods by a notable margin. The best and the second-best results are highlighted in \textcolor{red}{red} and \textcolor{blue}{blue}, respectively. Best viewed in zoom.}
\label{tab:quantComp}
\end{table*}


%% file: tables/od_cross_eval.tex
\begin{table}[!htb]
    \rowcolors{1}{gray!15}{white}
        \centering
        \scalebox{0.43}{\begin{tabular}{cccccccccccccc}
        \toprule
        Dataset/Class & Bicycle & Boat  & Bottle & Bus   & Car   & Cat   & Chair & Cup   & Dog   & Motorbike & People & Table & Avg.   \\ \midrule
        ExDark \cite{loh2019getting}       & 43.10   & 27.50 & 28.30  & 60.70 & 42.70 & 25.60 & 21.90 & \textcolor{blue}{32.30} & 29.60 & 21.40     & 29.00  & 21.70 & 32.00 \\
        LOL-V1 \cite{wei2018deep}       & \textcolor{blue}{56.00}   & 29.70 & \textcolor{blue}{35.50}  & \textcolor{blue}{63.50} & \textcolor{blue}{46.70} & 36.90 & 31.10 & 30.60 & \textcolor{blue}{38.80} & \textcolor{blue}{23.40}     & \textcolor{blue}{35.60}  & 27.50 & \textcolor{blue}{37.90} \\
        LOL-V2 \cite{yang2021sparse}        & 48.40   & 27.00 & 27.50  & 60.20 & 40.60 & 33.30 & 31.20 & 27.10 & 35.30 & 19.50     & 32.20  & \textcolor{red}{30.20} & 34.40 \\
        LSRW   \cite{hai2023r2rnet}          & \textcolor{red}{59.10}   & \textcolor{red}{31.80} & \textcolor{red}{36.50}  & \textcolor{red}{63.60} & 43.90 & 34.40 & 29.00 & 23.50 & 31.70 & 16.40     & 32.50  & 29.10 & 35.90 \\
        NCLLIE \cite{liu2024ntire}       & 51.90   & \textcolor{blue}{31.40} & 33.60  & 62.60 & 44.20 & \textcolor{blue}{38.50} & \textcolor{blue}{31.70} & 29.50 & 34.00 & 20.80     & 33.90  & \textcolor{blue}{30.00} & 36.80 \\ \midrule
        LSD           & 54.00   & 31.00 & 35.30  & 61.80 & \textcolor{red}{47.80} & \textcolor{red}{40.50} & \textcolor{red}{33.10} & \textcolor{red}{32.50} & \textcolor{red}{39.70} & \textcolor{red}{24.20}     & \textcolor{red}{36.20}  & \textcolor{blue}{30.00} & \textcolor{red}{38.80} \\ \bottomrule
        \end{tabular}}
         \caption{Object-detection on low-light scenarios. The proposed LSD can enhance the visibility and details of low-light images, resulting in significant detection improvement. The best and the second-best results are highlighted in \textcolor{red}{red} and \textcolor{blue}{blue}, respectively. }
  \label{tab:odcom}
\vspace{-.1cm}
\end{table}

%% file: tables/abl_lsd.tex
\begin{table}[!htb]
\centering
\rowcolors{1}{gray!15}{white}
\scalebox{0.5}{\begin{tabular}{cccccccccc}
\toprule
\multirow{2}{*}{Scene Type} & \multicolumn{3}{c}{DLI}                   & \multicolumn{3}{c}{NLI}                   & \multicolumn{3}{c}{LSD (Average)}                   \\ \cmidrule(lr){2-4} \cmidrule(lr){5-7} \cmidrule(lr){8-10} 
                                     & PSNR $\uparrow$  & SSIM $\uparrow$  & LPIPS $\downarrow$ & PSNR $\uparrow$  & SSIM $\uparrow$  & LPIPS $\downarrow$ & PSNR $\uparrow$  & SSIM $\uparrow$  & LPIPS $\downarrow$ \\ \midrule
Raw Patch                              & 15.47          & 0.6330          & 0.3107          & 15.80          & \textcolor{blue}{0.6293}          & \textcolor{blue}{0.3155}          & 15.64          & \textcolor{blue}{0.6312}          & 0.3131          \\
Full-scene                                 & \textcolor{blue}{17.52}          & \textcolor{blue}{0.6396}          & \textcolor{blue}{0.2770}          & \textcolor{blue}{17.09}          & 0.6095          & 0.3239          & \textcolor{blue}{17.31}          & 0.6246          & \textcolor{blue}{0.3005}          \\ \hline
\textbf{Filtered Patch}                   & \textcolor{red}{19.13} & \textcolor{red}{0.7091} & \textcolor{red}{0.2609} & \textcolor{red}{18.74} & \textcolor{red}{0.6995} & \textcolor{red}{0.2661} & \textcolor{red}{18.93} & \textcolor{red}{0.7043} & \textcolor{red}{0.2635}   \\ \bottomrule
\end{tabular}}
\caption{Ablation study on LSD. The proposed filtered patches help the deep model to produce more natural-looking enhanced images. The best and the second-best results are highlighted in \textcolor{red}{red} and \textcolor{blue}{blue}. }
  \label{tab:abllsd}

\end{table}

%% file: tables/abl_tf.tex
\begin{table}[!htb]
\centering
\rowcolors{1}{gray!15}{white}
\scalebox{0.43}{\begin{tabular}{ccccccccccccc}
\toprule
\multicolumn{4}{c}{Module}                               & \multicolumn{3}{c}{DLI}                   & \multicolumn{3}{c}{NLI}                 & \multicolumn{3}{c}{LSD (Average)}                   \\ \cmidrule(lr){5-7} \cmidrule(lr){8-10}\cmidrule(lr){11-13}  
LCE & LCCAB & LCGRB & LCCG & PSNR $\uparrow$  & SSIM $\uparrow$  & LPIPS $\downarrow$ & PSNR $\uparrow$  & SSIM $\uparrow$ & LPIPS $\downarrow$& PSNR $\uparrow$  & SSIM $\uparrow$  & LPIPS $\downarrow$ \\ \midrule
             \textcolor{red}{\xmark }     &         \textcolor{red}{\xmark }     &       \textcolor{red}{\xmark }       &   \textcolor{red}{\xmark }                & 15.46          & 0.6248          & 0.2856          & 15.27          & 0.5942         & 0.2949         & 15.36          & 0.6095          & 0.2903          \\
                \textcolor{green}{\cmark }&      \textcolor{red}{\xmark }          &      \textcolor{red}{\xmark }        &  \textcolor{red}{\xmark }                  & 16.65          & 0.6469          & 0.2817          & 16.82          & 0.6265         & {0.2775}         & 16.74          & 0.6367          & 0.2796          \\ \textcolor{green}{\cmark }
                 &       \textcolor{green}{\cmark }       &      \textcolor{red}{\xmark }      &     \textcolor{red}{\xmark }            & 17.66          & 0.6805          & 0.2783          & 17.87          & 0.6506         & \textcolor{blue}{0.2728}         & 17.77          & 0.6656          & \textcolor{blue}{0.2756}          \\ \textcolor{green}{\cmark }
                 &         \textcolor{green}{\cmark }     &   \textcolor{green}{\cmark }         &     \textcolor{red}{\xmark }            & \textcolor{blue}{17.74}          & \textcolor{blue}{0.6856}          & \textcolor{blue}{0.2706}          & \textcolor{blue}{17.99}          & \textcolor{blue}{0.6697}         & 0.2840          & \textcolor{blue}{17.86}          & \textcolor{blue}{0.6776}          & 0.2773          \\ \midrule \textcolor{green}{\cmark }
                 &      \textcolor{green}{\cmark }        &     \textcolor{green}{\cmark }       &   \textcolor{green}{\cmark }             & \textcolor{red}{19.13} & \textcolor{red}{0.7091} & \textcolor{red}{0.2609} & \textcolor{red}{18.74} & \textcolor{red}{0.6995} & \textcolor{red}{0.2661} & \textcolor{red}{18.93} & \textcolor{red}{0.7043} & \textcolor{red}{0.2635} \\ \bottomrule
\end{tabular}}
\caption{Ablation study on TFFormer. The proposed component helps the TFFormer in achieving SOTA performance on LSD. The best and the second-best results are highlighted in \textcolor{red}{red} and \textcolor{blue}{blue}.}
  \label{tab:abltff}
\end{table}

%% file: sec/5_conclusion.tex
\section{Conclusion}

We presented LSD, the largest real-world SLLIE dataset to date, comprising 6,425 high-resolution low-light image pairs captured using diverse devices and adaptive calibration strategies. From these, 6,025 pairs were used to generate approximately 180,000 training patches, with 400 full-resolution pairs reserved for benchmarking. Additionally, we collected 2,117 unpaired low-light images from varied sources to support aesthetic and generalization evaluations. To fully leverage this dataset, we proposed a novel Transformer-CNN model that decouples luminance and chrominance features for accurate color restoration and robust denoising. Our design integrates LCCAB, LCGRB, and LCPL modules to enhance global fusion, local refinement, and perceptual quality. Extensive evaluations demonstrate strong generalization across hardware, with future extensions aimed at joint SLLIE–object detection and extreme NLI enhancement.

%% file: sec/acknowledgement.tex
\section{Acknowledgment}
This research was supported in part by the Alexander von Humboldt Foundation.

%% file: sec/sup.tex

This supplementary document details the LSD preparation and insights of TFFormer, provides additional experimental results, and discusses the limitations of the proposed method. We organized the document as follows:
\begin{itemize}
    \item Section \ref{lsdpre} details LSD preparation
    \item Section \ref{sec:lsddistsllie} details the distribution of the LSD and illustrates the limitation of the existing dataset visually.
    \item Section \ref{tfformerdet} provides learning details and performance of the TFFormer in existing SLLIE.
    \item Section \ref{extrares} provides additional experimental results
\end{itemize}

\section{LSD Preparation}
\label{lsdpre}
We designed our LSD app carefully to collect a large-scale SLLIE dataset leveraging multiple sensors for diverse real-world data collection. We also developed our scene classifier by utilizing the capability of SOTA image classification models.

\subsection{Sensor Details}
To diversify our proposed LSD dataset, we utilize 15 distinct image sensors from six different smartphones, encompassing a variety of imaging principles, resolutions, pixel sizes, and sensor sizes. These sensors respond uniquely to various lighting conditions, particularly noise, dynamic range, and color reproduction. Table \ref{tab:camera_specs} provides the specifications of these smartphones and their cameras, with the resolution reflecting the maximum native output of each sensor. However, under low-light conditions, these sensors typically generate 4K images by leveraging pixel-binning techniques to create larger effective pixels, enhancing brightness and reducing noise.

This study replicates real-world scenarios using the default pixel-binning configurations of modern smartphones. Only the rear cameras were employed, as they offer superior imaging performance compared to front-facing cameras. The diversity in sensor types, field of view, and imaging characteristics substantially enriches the LSD dataset, ensuring its robustness and relevance for low-light image enhancement research.

\begin{table*}[!htb]
\centering
\rowcolors{1}{gray!15}{white}
\scalebox{0.8}{\begin{tabular}{cccccccc}
\hline
\textbf{Device} & \textbf{Camera}       & \textbf{Sensor} & \textbf{ Resolution (Max.)} & \textbf{Sensor Size} & \textbf{Pixel Size} & \textbf{Aperture} & \textbf{Year} \\ \hline
Samsung Galaxy S10    & Main (Wide)     & S5K2L4       & 12 MP (4032 x 3024)             & 1/2.55 in.        & 1.4 µm              & f/1.5-2.4         & 2019          \\ \cline{2-8} 
                      & Telephoto       & S5K3M3       & 12 MP (4032 x 3024)             & 1/3.6 in.         & 1.0 µm              & f/2.4             & 2019          \\ \cline{2-8} 
                      & Ultra-Wide      & S5K4H5       & 16 MP (4608 x 3456)             & 1/3.1 in.         & 1.0 µm              & f/2.2             & 2019          \\ \hline
Samsung Galaxy Z Flip3 & Main (Wide)    & IMX555  & 12 MP (4032 x 3024)             & 1/2.55 in.        & 1.4 µm              & f/1.8             & 2021          \\ \cline{2-8} 
                      & Ultra-Wide      & IMX258 & 12 MP (4000 x 3000)             & 1/3.0 in.         & 1.12 µm             & f/2.2             & 2021          \\ \hline
Xiaomi Redmi 10C       & Main (Wide)    & ISOCELL JN1  & 50 MP (8160 x 6144)             & 1/2.76 in.        & 0.64 µm             & f/1.8             & 2022            \\ \hline
Samsung Galaxy S22 Ultra & Main (Wide)  & HM3          & 108 MP (12032 x 9024)           & 1/1.33 in.        & 0.8 µm              & f/1.8             & 2022          \\ \cline{2-8} 
                      & Telephoto (3x)  & IMX754       & 10 MP (3648 x 2736)             & 1/3.52 in.        & 1.12 µm             & f/2.4             & 2022          \\ \cline{2-8} 
                      & Telephoto (10x) & IMX754       & 10 MP (3648 x 2736)             & 1/3.52 in.        & 1.12 µm             & f/4.9             & 2022          \\ \cline{2-8} 
                      & Ultra-Wide      & IMX563       & 12 MP (4000 x 3000)             & 1/2.55 in.        & 1.4 µm              & f/2.2             & 2022          \\ \hline
Samsung Galaxy Z Flip5 & Main (Wide)    & IMX563       & 12 MP (4032 x 3024)             & 1/1.76 in.        & 1.8 µm              & f/1.8             & 2023          \\ \cline{2-8} 
                      & Ultra-Wide      & IMX258       & 12 MP (4000 x 3000)             & 1/2.55 in.        & 1.12 µm             & f/2.2             & 2023          \\ \hline
Samsung Galaxy Z Fold5 & Main (Wide)    & GN5          & 50 MP (8160 x 6120)             & 1/1.57 in.        & 1.0 µm              & f/1.8             & 2023          \\ \cline{2-8} 
                      & Telephoto       & IMX754       & 10 MP (3648 x 2736)             & 1/3.94 in.        & 1.0 µm              & f/2.4             & 2023          \\ \cline{2-8} 
                      & Ultra-Wide      & IMX563       & 12 MP (4000 x 3000)             & 1/3.0 in.         & 1.12 µm             & f/2.2             & 2023          \\ \hline
\end{tabular}}
\caption{Specifications of the camera sensors utilized to collect the proposed LSD. Utilizing such diverse sensors helps us generalize the SLLIE in the real world for practical use.}
\label{tab:camera_specs}
\end{table*}

\subsection{LSD App}
\begin{figure}[!htb]
  \centering
  \includegraphics[width=.85\linewidth]{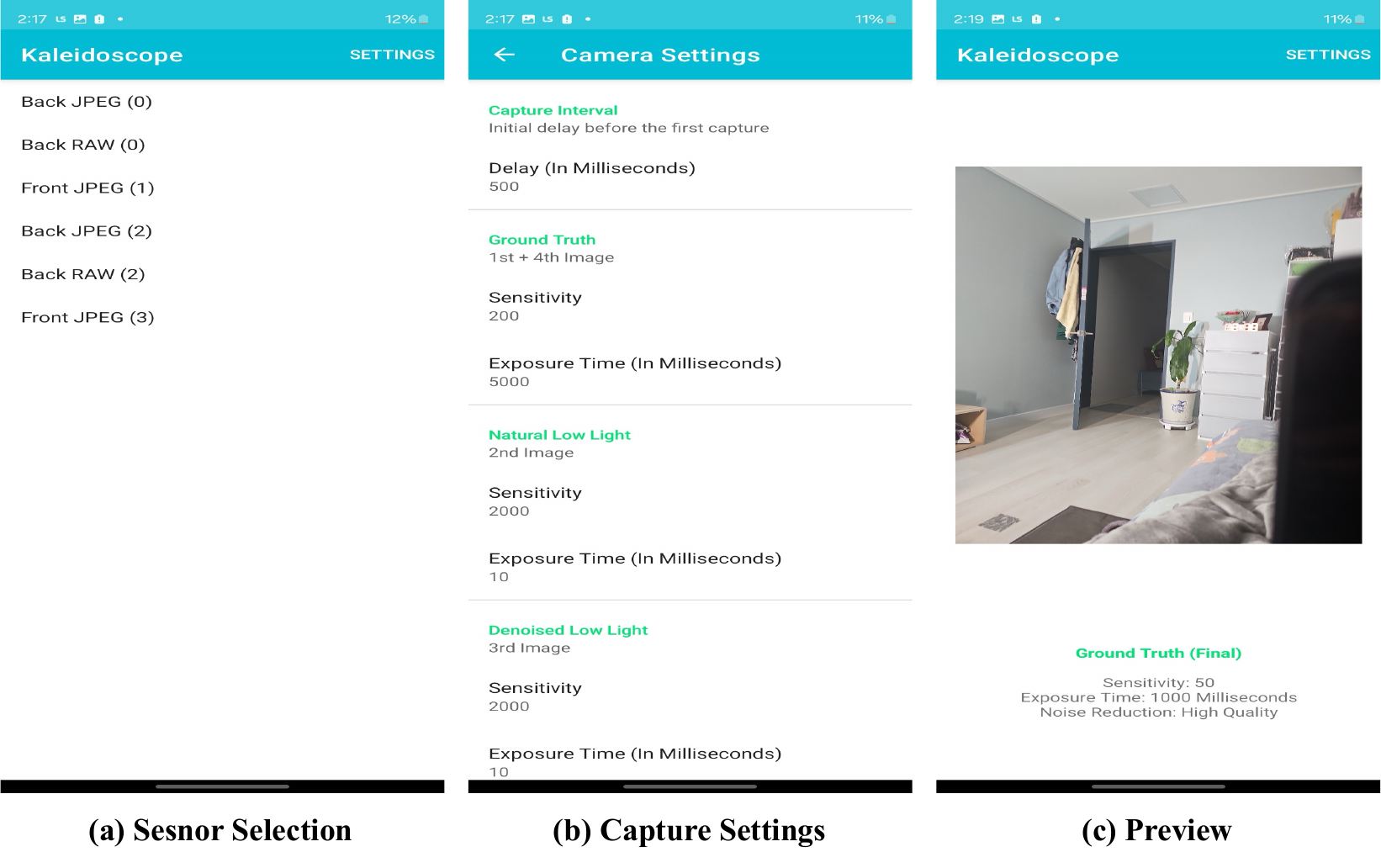}
  \caption{User interface and detail of LSD APP. (a) Selecting Sensor. (b) Settings adjustment of low-light and reference images. (c) Preview of capture scenes}
  \label{fig:lsdapp}
\end{figure}

Our LSD app was developed by leveraging android-api \cite{android-camera-api} to capture aligned low-light and corresponding reference images. Before fixing the final version, we experimented with the available APIs to find the best and easy-to-use interface for collecting low-light samples. Fig. \ref{fig:lsdapp} illustrates the sample screenshot of our LSD app. To capture a scene, we first selected a random camera sensor for the device; we calibrated the settings based on our capture calibration strategy. Before saving the scenes, we visually inspected the captured scenes to ensure their usability. 

\subsection{VGGScene Classifier}
Refining the collected patches from our image pairs posed significant challenges, mainly due to unwanted imaging limitations such as blurs, over-exposed light sources, dark under-exposed regions, and defocus in reference images. Developing individual algorithms to address each issue proved inefficient and complex, while manually inspecting every image pair was impractical.

To overcome these hurdles, we developed a learning-based scene classifier to automate and accelerate the refinement process. This approach enabled us to identify and exclude imperfect images from training, ensuring higher dataset quality. Before beginning the LSD scene collection, we extensively evaluated state-of-the-art (SOTA) classification methods and constructed a separate scene classification dataset to fine-tune our classifier. This allowed us to handle challenging scenarios and streamline the refinement process robustly.

\subsubsection{Data Preparation.}
\begin{figure*}[!htb]
  \centering
  \includegraphics[width=\linewidth,height=3.2cm]{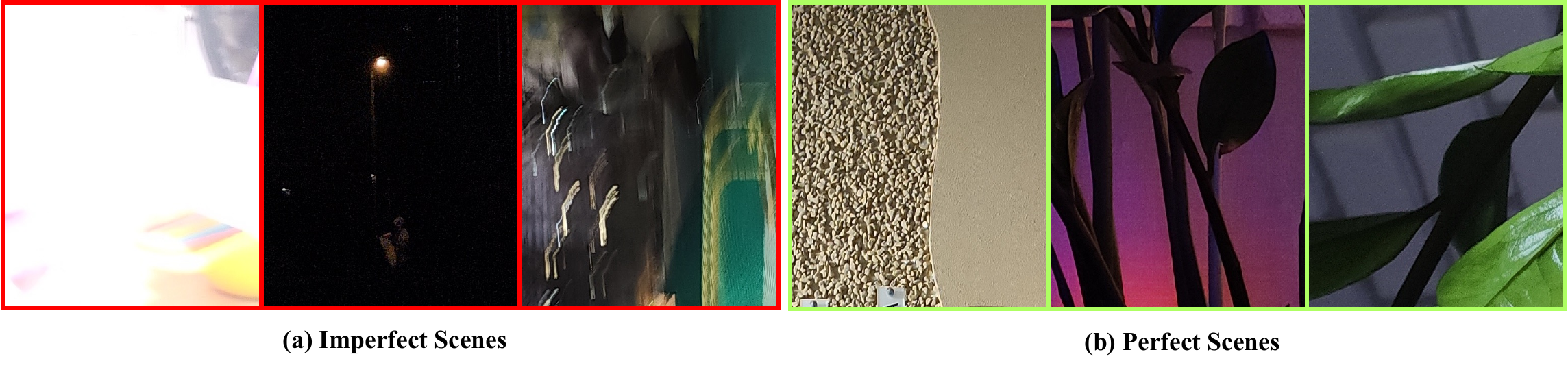}
  \caption{Sample images from scene classification dataset. (a) Imperfect scenes. (b) Perfect scenes.}
  \label{fig:sceneclass}
\end{figure*}

Before start collecting the LSD dataset, we captured around 200 random scenes to find the limitations and probable obstacles of preprocessing our LSD. Also, we made a list of unwanted imaging consequences that may affect our learning strategy. Based on the initial study, we made a classification dataset and separated perfect and imperfect scenes into two categories. Tab. \ref{tab:classcomp} illustrates the detail of our scene classification dataset. We collected 12,770 image patches to learn scene classifiers with SOTA classification methods. Fig. \ref{fig:sceneclass} demonstrates the sample images from our scene classification dataset.

\begin{table}[!htbt]
\centering
\rowcolors{1}{gray!15}{white}
\begin{tabular}{ccc} \toprule
Class          & Train          & Test          \\ \midrule
Perfect        & 5,580           & 755           \\
Imperfect      & 5,456           & 979           \\ \midrule
\textbf{Total} & \textbf{11,036} & \textbf{1,734} \\ \bottomrule
\end{tabular}

\caption{ Detail of LSD scene classification dataset.}
\label{tab:classcomp}
\end{table}

\subsubsection{Experiments and Results.} 
\begin{figure}[!htb]
  \centering
  \includegraphics[width=\linewidth, height=3.4cm]{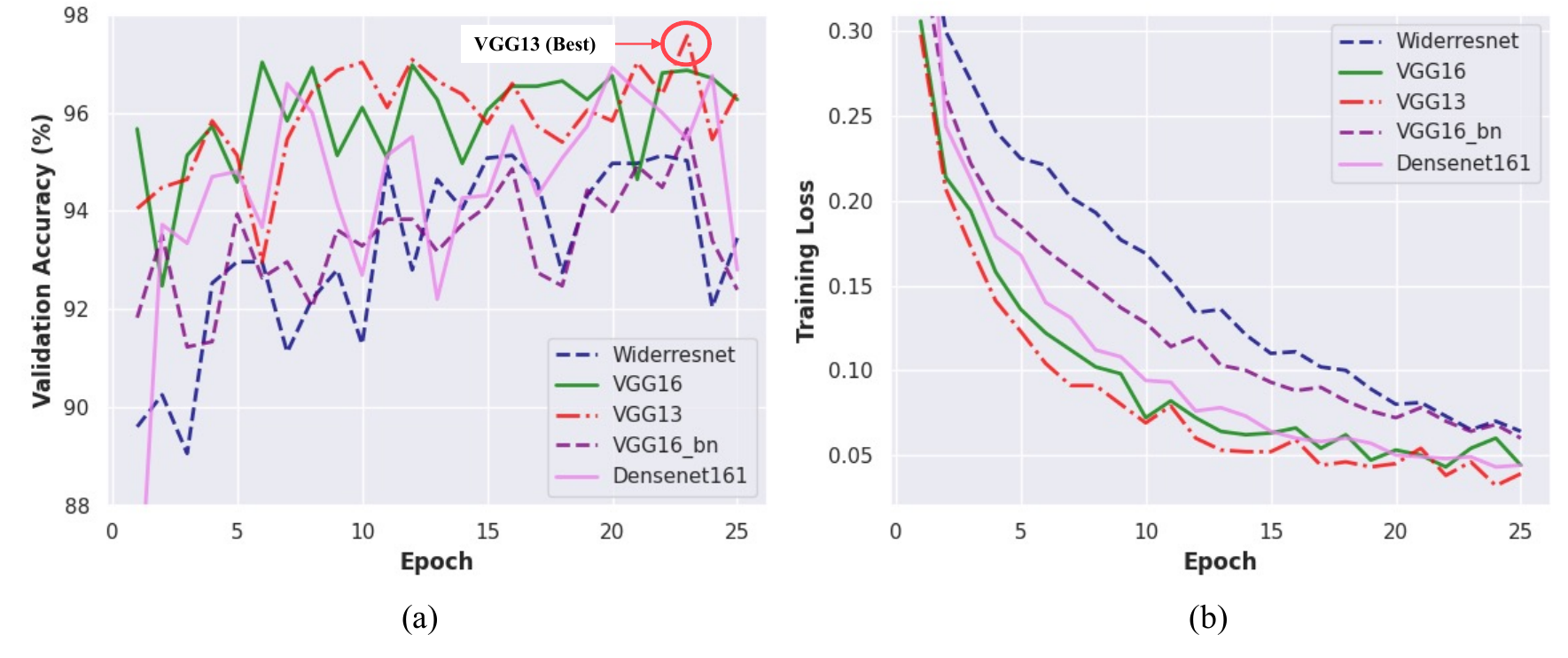}
  \caption{Learning scene classification with SOTA image classification methods. (a) Validation accuracy vs epoch. (b) Training loss vs epoch.}
  \label{fig:vgglogs}
\end{figure}

We studied the existing classification method to learn the perfect scene identification. We alter the final classification layer of SOTA models to fit our objective \cite{subramanian2018deep}. Additionally, we leverage Imagenet's \cite{deng2009imagenet} pre-trained weights of these methods from the torch-vision \cite{pytorch-pretrained} library to accelerate the learning process and achieve faster convergence. We trained existing methods for 25 epochs with their suggested hyperparameters. Fig. \ref{fig:vgglogs} details the training loss and validation accuracy of the SOTA classification method on LSD scene classification. VGG13 \cite{liu2015very} illustrates the maximum validation accuracy among the SOTA classification methods. Based on the experimental results, we leverage the best weight of VGG13 for making our LSD scene classifier. Tab. \ref{tab:vggclass} compares numerous deep methods in LSD scene classification.

\begin{table}[!htb]
\centering
\rowcolors{1}{gray!15}{white}

\begin{tabular}{cc} \toprule
Model       & Accuracy (\%)   \\ \midrule
Widerresnet \cite{wu2019wider} & 95.13           \\
VGG16  \cite{liu2015very}     & 97.02          \\
VGG16\_bn  \cite{liu2015very}  & 95.67          \\
Densenet161 \cite{huang2017densely} & 96.92          \\
VGG13   \cite{liu2015very}    & \textbf{97.57} \\ \bottomrule
\end{tabular}

\caption{Comparison between SOTA deep image classification methods on LSD scene classification.}
\label{tab:vggclass}
\end{table}

\begin{figure}[!htb]
    \centering

  \includegraphics[width=.98\linewidth, height=4.5cm]{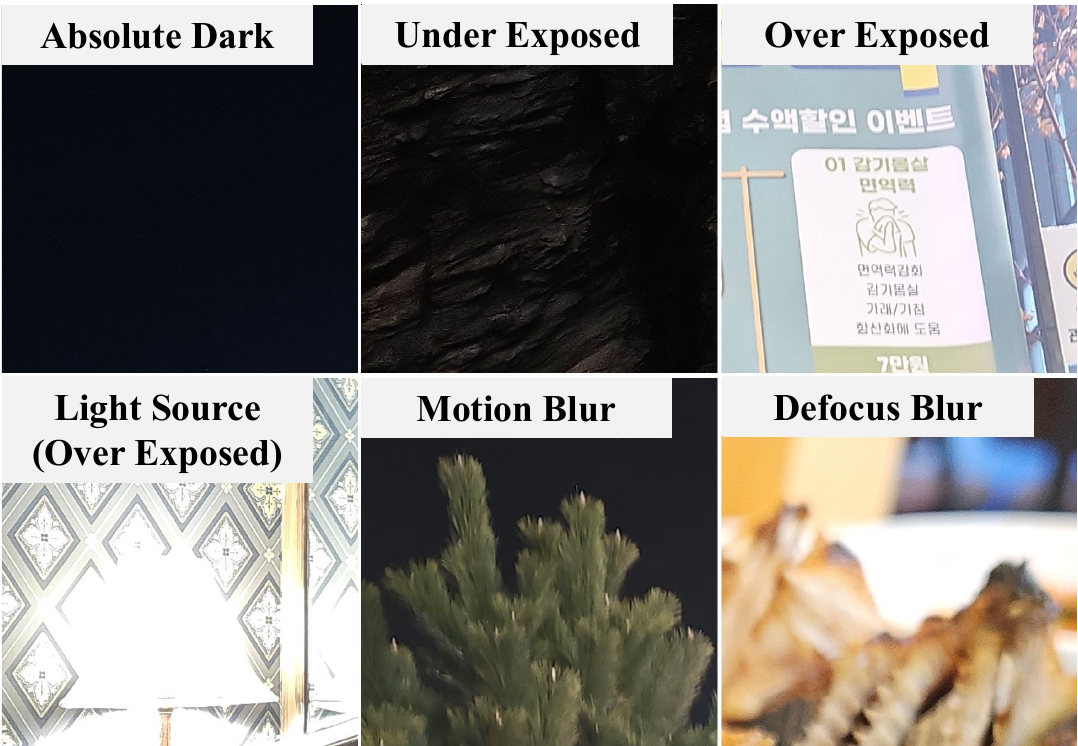}
 
  \caption{Sample images were eliminated through our  patch filtering  strategy.}
  \label{fig:elim}

\end{figure}

Fig. \ref{fig:elim} illustrates the sample images eliminated by our patch filtering strategy.

\section{LSD vs. Existing Datasets}
\label{sec:lsddistsllie}
\subsection{Training Scenes}
Capturing images in real-world, uncontrolled environments offers significant advantages, particularly in replicating scenarios where images are taken with handheld cameras. However, such conditions often introduce imaging limitations, such as blurs, over-exposed highlights, and under-exposed regions, especially in the reference images. These imperfections can mislead deep learning models and result in unusable outputs.

To address this, we refined our training dataset by filtering out images with these limitations. This was achieved by analyzing the global intensity of reference images and leveraging our VGG-based scene classifier for robust filtering. Tab .\ref{tab:traindist} provides a detailed breakdown of the device-wise distribution of the refined LSD training dataset, highlighting the diversity and quality ensured through this process. It is worth noting that some devices may use the same image sensor. However, these setups employ different focal lengths and image processing techniques, ensuring that the resulting samples remain diverse even when the same sensor is utilized.

\begin{table*}[!htb]
\centering
\rowcolors{1}{gray!15}{white}
\begin{tabular}{ccccccc}
\toprule
Device         & DLI           & NLI           & Combine       & Raw Patches     & Filtered Patches & Usable Patch (\%) \\ \midrule
Samsung Galaxy S10            & 84            & 67            & 151           & 5,185            & 3,847             & 74.19             \\
Samsung Galaxy Z Fold 5          & 759           & 689           & 1448          & 56,577           & 50,416            & 89.11             \\
Samsung Galaxy S 22 Ultra       & 1,270          & 1,420          & 2,690          & 85,471           & 75,514            & 88.35             \\
Samsung Galaxy Z Flip 3          & 329           & 347           & 676           & 22,800           & 20,312            & 89.09             \\
Xiaomi Redmi10C       & 166           & 161           & 327           & 9,163            & 8,439             & 92.10             \\
Samsung Galaxy Z Flip5          & 367           & 366           & 733           & 29,657           & 21,167            & 71.37             \\ \bottomrule
\textbf{Total} & \textbf{2,975} & \textbf{3,050} & \textbf{6,025} & \textbf{208,853} & \textbf{179,695}  & \textbf{86.04}    \\ \bottomrule
\end{tabular}

\caption{Details of LSD training set (filtered). We filtered out around 15\% of patch pairs to make our training set robust.  }
\label{tab:traindist}
\end{table*}

\subsection{Testing Scenes}
One major limitation of the current SLLIE datasets is the absence of diverse testing scenes. To counter this, we curated the most extensive benchmarking test set, incorporating a variety of hardware, scenes, and sources. Tab. \ref{tab:pairs} details our LSD test set. For comprehensive evaluation, our test set includes paired scenes for quantitative and qualitative assessments and unpaired images for subjective evaluation.

\subsubsection{Pair Images.} We collected 400 testing scenes for pair testing, including DLL and NLL pairs captured in different lighting conditions. Each of these categories also includes subsets for indoor and outdoor samples. For testing scenes, we selected the full scenes with minimal unexpected imaging instances. 

\subsubsection{Unpair Images.} Image enhancement invariably remains debatable, contingent upon personal preferences. Therefore, the proposed LSD offers 2,117 unique test cases for extensive subjective assessment. Our unpaired testing set encompasses data samples captured with DSLR cameras, smartphones, frames from 10 low-light videos, and social media images. We observed millions of images stored on social media, which can still benefit from SLLIE techniques, breathing new life into them.

\begin{table}[!htb]
    \centering
    \rowcolors{1}{gray!15}{white}
\scalebox{0.9}{\begin{tabular}{cccc}
\toprule
Type               & Lighting Condition   & Category                  & Scenes        \\ \midrule
   & Low-light                  & DLI                       & 100           \\
                 Pair       & Extreme              & DLI                       & 100           \\
                        & Low-light                  & NLI                       & 100           \\
                        & Extreme              & NLI                       & 100           \\ \midrule  &  & Android                   & 300           \\
                        &                      & iPhone                    & 250           \\
                 Unpair     &          Mix              & DSLR                      & 300           \\
                        &                      & Social Media              & 50            \\
                        &                      & Video Frames & 1,217          \\ \midrule
\multicolumn{3}{c}{\textbf{Total}}                                                  & \textbf{2,517} \\ \bottomrule
\end{tabular}}

\caption{Detail of our LSD benchmarking set. We developed the largest and most diverse SLLIE benchmarking dataset throughout this study.}
\label{tab:pairs}
\end{table}

\subsection{Limitation of Existing SLLIE Datasets}

\begin{figure*}[!htb]
  \centering
  \includegraphics[width=\linewidth]{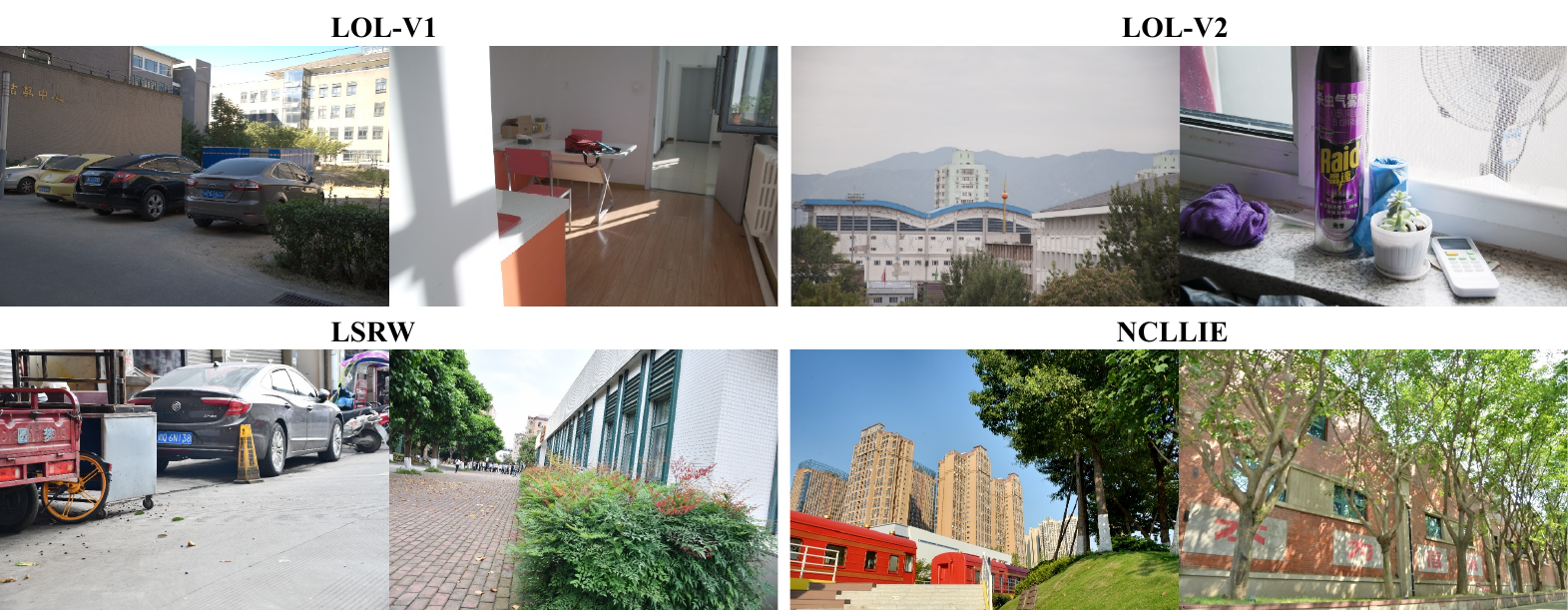}
  \caption{Sample images from existing SLLIE dataset. Many of the collected scenes of these datasets are captured in bright outdoor scenes and replicated in the low-light input by controlling camera parameters. }
  \label{fig:sllieD}
\end{figure*}

Existing SLLIE datasets predominantly consist of scenes captured under well-lit conditions, with low-light inputs artificially generated using low ISO settings and short exposure times. As shown in Fig. \ref{fig:sllieD}, these samples are often taken in high-brightness outdoor environments, making the dataset semi-synthesized rather than representative of actual low-light scenarios. While useful for controlled experiments, this approach fails to authentically replicate the challenges faced in natural low-light environments, such as complex lighting distributions, varying noise levels, and diverse low-illumination conditions.

Moreover, these brighter scenes often contain naturally underexposed regions better suited for HDR mapping than developing and evaluating dedicated SLLIE methods. This mismatch can mislead model development, as the dataset does not fully capture the nuanced challenges of real low-light imaging. To advance SLLIE research, it is essential to focus on datasets that authentically represent diverse low-light conditions rather than relying on semi-synthetic approximations.

\begin{figure*}[!htb]
  \centering
  \includegraphics[width=\linewidth]{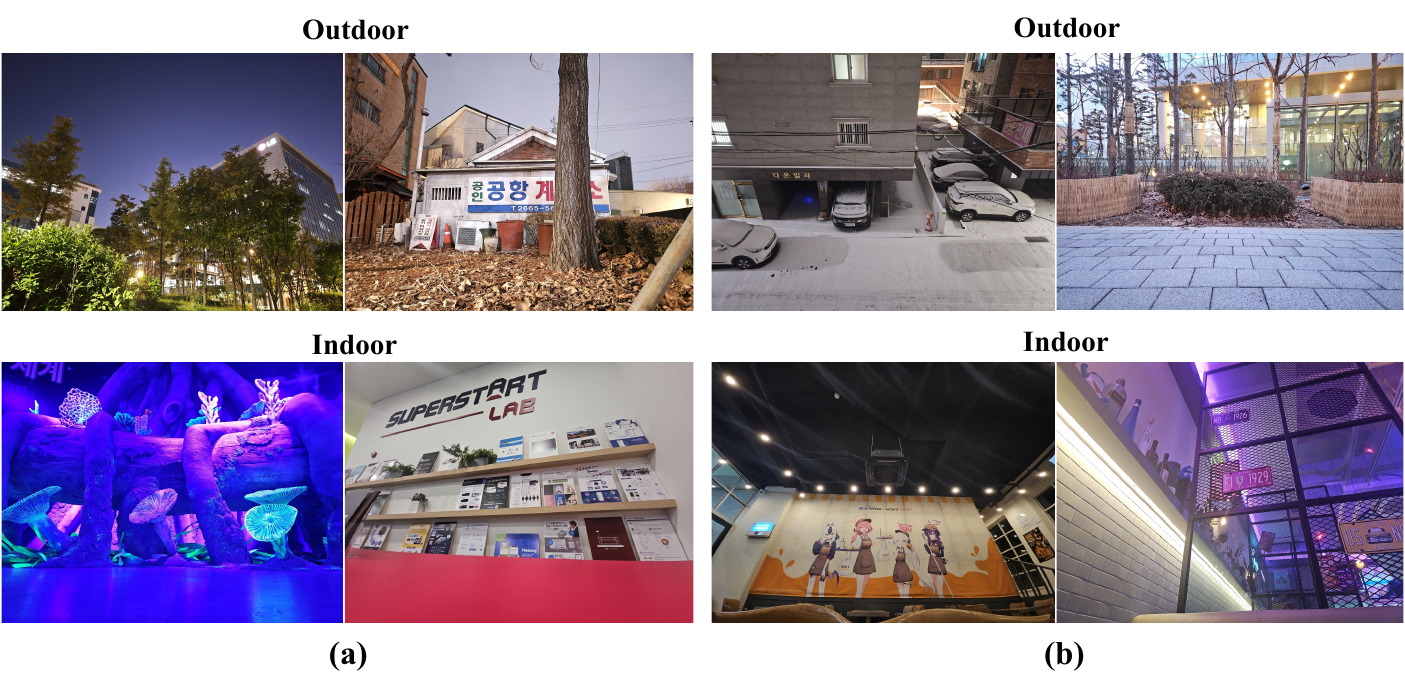}
  \caption{Sample images LSD. We collected all samples in 0.1-200 lux for over two years. (a) Samples from DLI scenes. (b) Samples from NLI scenes}
  \label{fig:lsddiverse}
\end{figure*}

In contrast to the existing SLLIE datasets, we collected our datasets in actual low-light scenarios (0.1-200 lux). Fig. \ref{fig:lsddiverse} illustrates the samples from the proposed LSD. We collected data over the years in different seasons, such as winter, autumn, and summer.

\section{TFFormer}
\label{tfformerdet}
Due to page limits, our main article could not provide details on implementing TFFormer, LC loss, and complexity analysis. This section details the missing part to ensure the reproducibility of our proposed method. 

\subsection{Implementation Details}

\textbf{LC Extraction:} Luminance-Chrominance (LC) extraction separates an image's intensity and color information. Luminance (L) is computed as L = 0.299R + 0.587G + 0.114B, representing brightness based on human visual perception. Where R, G, and B are color channels of an image. Chrominance (C) captures color details by subtracting luminance from the original Image: C = I - L. This decomposition is widely used in image processing for tasks like compression, enhancement, and object tracking, enabling better independent handling of brightness and color. \newline 

\textbf{LC Encoding:} We expanded the luminance-boosted image ($\mathbf{I_{BL}}$), luminance features ($\mathbf{F_L}$), chrominance-boosted image ($\mathbf{I_{BC}}$) and chrominance features ($\mathbf{F_C}$) into same feature dimension as $\mathbf{F_{x}} \in \mathbb{R}^{H \times W \times 40}$. Later, we feed our luminance and chrominance encoder with these expanded feature maps to obtain $\mathbf{I_{L_{enc}}}$ or $\mathbf{I_{C_{enc}}}$. Our luminance and chrominance encoders comprise the LCGAB, followed by two convolution operations with stride 2 to down-sample the LC attributes and boosted-image maps. We expanded the dimension of the feature channel by a factor of 2 in every recurring encoder block. Also, we reduced the spatial dimension by a factor of 2  while expanding the feature maps in both encoders to obtain $\mathbf{I_{BL_{down}}}$ and $\mathbf{I_{FL_{down}}}$.

\begin{equation}
\mathbf{I_{L_{enc}}} = LCGAB(\mathbf{I_{BL}}, \mathbf{F_{L}})
\end{equation}
\vspace{-15pt}
\begin{equation}
\mathbf{I_{BL_{down}}} = Conv(\mathbf{I_{enc}}, s=2)
\end{equation}
\vspace{-15pt}
\begin{equation}
\mathbf{I_{FL_{down}}} = Conv(\mathbf{F_{L}}, s=2)
\end{equation}

The Chrominance branch follows the same equation to obtain their respective images $\mathbf{I_{C_{enc}}}$, $\mathbf{I_{BC_{down}}}$ and $\mathbf{I_{FC_{down}}}$.

\textbf{LC Decoding.} In the decoder part, we combined the luminance and chrominance features from the encoder by adding them as a skip connection to guide the decoder's LCGAB. The LCGAB in the decoder is followed by a transpose convolution operation to decode the RGB images into their actual input dimension. 

\begin{equation}
\mathbf{I_{L_{dec}}} = LCGAB(\mathbf{I_{BL_{down}}}, \mathbf{I_{FL_{down}}})
\end{equation}
\vspace{-15pt}
\begin{equation}
\mathbf{I_{UP}} = ConvTransposed(\mathbf{I_{L_{dec}}})
\end{equation}

Chrominance features are also up-sampled similarly. We refined the final decoded features with our reconstructed luminance and chrominance attributes. Additionally, our TFFormer is designed as a fully convolutional network and can take images of any dimension as input. It can generate the output with the same dimension as the input without losing any spatial dimension.  

\
\subsection{Loss calculation}
 To perceive visually plausible images, we leverage an additional L1 loss along with the proposed LCCG. Thereby, we defined final LC-guided loss as:

\begin{equation}
\mathcal{L} = \lambda_R (\mathcal{L}_R + \mathcal{L}_{LC})
\end{equation}
where $\mathcal{L}_R$ represents the traditional $L_1$ loss, $\lambda_R$ is a low-light regularization coefficient. 



\subsection{Training Details}
We implemented our proposed TFFormer using PyTorch framework \cite{pytorch}. Our TFFormer optimized with an  Adam optimizer \cite{kingma2014adam} with a hyperparameter of $\beta_1 = 0.9$, $\beta_2 = 0.99$. We set the initial learning rate = $1e-4$ and adjusted it after training 100,000 steps using \texttt{ReduceLROnPlateau} scheduler \cite{pytorch-docs}. We trained our model for 250,000 steps, utilizing a batch size of 12 and image dimensions $128\times128\times3$. Additionally, we leverage a low-light regularization coefficient $\lambda_R = 0.2$ empharically to tune the proposed LC guidance. The training process spanned approximately 100 hours, executed on low-end hardware featuring an AMD Ryzen 3200G central processing unit (CPU) operating at 3.6 GHz, complemented by 32 GB of random-access memory, and an Nvidia GeForce GTX 3060 (12GB) graphical processing unit (GPU).

\subsection{TFFormer on SLLIE Datasets}
To assess the performance of the proposed LSD dataset against existing SLLIE datasets, we leverage our TFFormer model throughout this study. Prior to conducting cross-dataset comparisons, we validated the effectiveness of TFFormer on existing SLLIE datasets to ensure a fair and consistent evaluation. For this purpose, we compared TFFormer with state-of-the-art (SOTA) SLLIE methods that utilize Retinex theory or transformer-based architectures. Tab. \ref{tab:lolQuant} summarizes these comparisons, focusing on methods like Retinexformer, recognized as one of the best single-stage SLLIE models and the winner of the NCLLIE-CVPR24 challenge. Additionally, we included only models for which reported results were reproducible.

As shown in Tab. \ref{tab:lolQuant}, TFFormer consistently outperformed all SOTA methods across LOL-V1 \cite{wei2018deep} and LOL-V2 (real) \cite{yang2021sparse}  datasets. Specifically, TFFormer achieved the highest PSNR (26.13 dB on LOL-V1 and 31.55 dB on LOL-V2), SSIM (0.8875 on LOL-V1 and 0.9147 on LOL-V2), and the lowest LPIPS (6.12 and 3.79, respectively). These results highlight the superior performance of TFFormer in enhancing real-world low-light images, surpassing strong baselines such as Retinexformer and MIRNet. This establishes TFFormer as a robust model for LSD-based evaluations and a strong contender for advancing SLLIE research on diverse datasets.

\begin{figure*}[!htb]
  \centering
  \includegraphics[width=\linewidth]{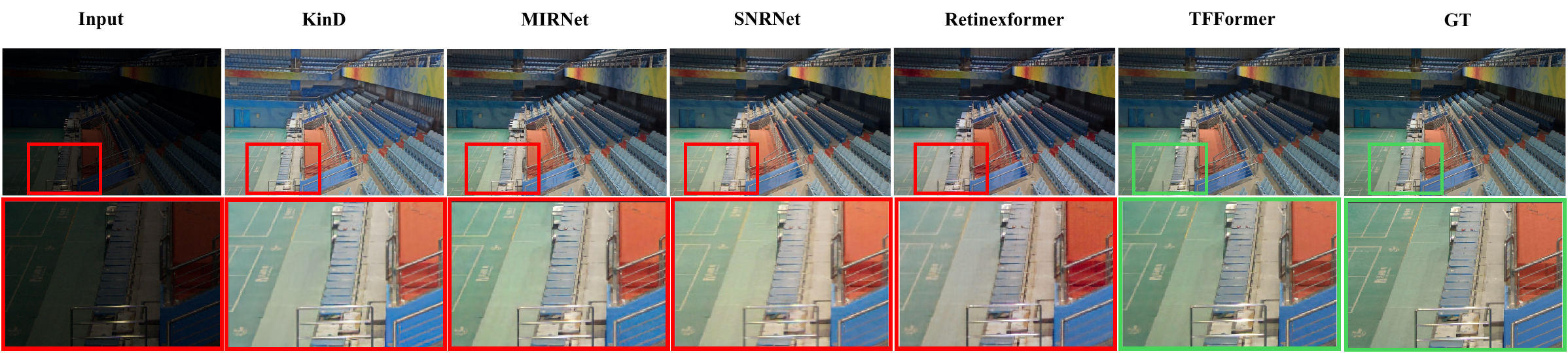}
  \caption{Qualatitive comparison between TFFormer and SOTA methods. The proposed method can produce cleaner and more color-accurate indoor scene images.}
  \label{fig:lolin}
\end{figure*}

\begin{figure*}[!htb]
  \centering
  \includegraphics[width=\linewidth]{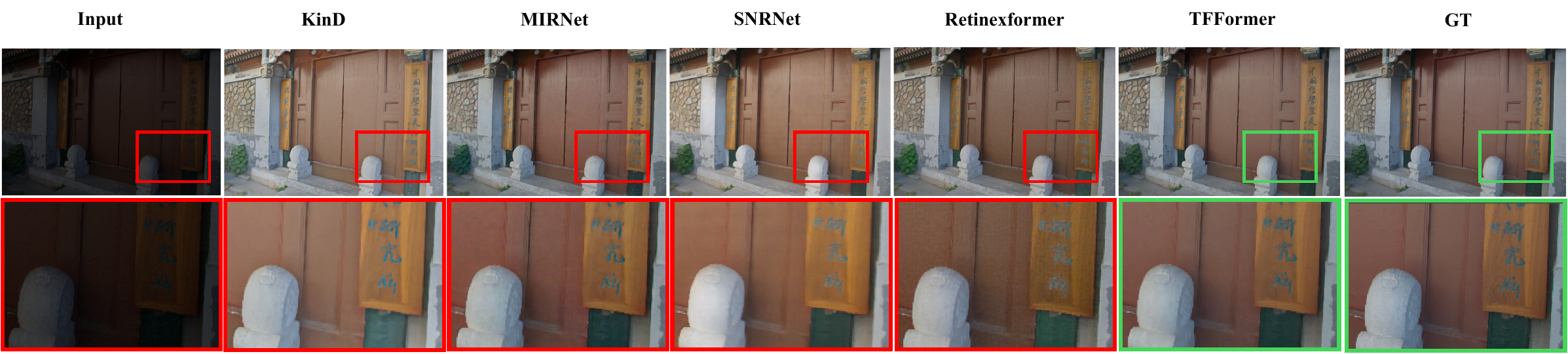}
  \caption{Qualatitive comparison between TFFormer and SOTA methods. The proposed method can produce cleaner and more color-accurate images on outdoor scenes.}
  \label{fig:lolout}
\end{figure*}

In addition to the quantitative evaluation, the qualitative results in Fig. \ref{fig:lolin} and Fig. \ref{fig:lolout} further demonstrate the practicability of TFFormer for generic SLLIE tasks. The proposed TFFormer consistently produces cleaner images with enhanced detail preservation and improved color accuracy, resembling the reference images. Additionally, the superior performance of TFFormer on LOL-V1 and LOL-V2 ensures its reliability and robustness for performing cross-dataset evaluations.

\begin{table*}[!htb]
\centering
\rowcolors{1}{gray!15}{white}
\begin{tabular}{cccccccccc}
\toprule
Method & \multicolumn{3}{c}{LOL-V1} & \multicolumn{3}{c}{LOL-V2} & \multicolumn{3}{c}{Average} \\ \cline{2-10}
                        & PSNR $\uparrow$    & SSIM $\uparrow$    & LPIPS $\downarrow$  & PSNR $\uparrow$    & SSIM $\uparrow$    & LPIPS $\downarrow$  & PSNR $\uparrow$    & SSIM $\uparrow$    & LPIPS $\downarrow$   \\ \midrule
Input                   & 7.77    & 0.4186  & 37.08  & 9.72    & 0.4370  & 27.24  & 8.75    & 0.4278  & 32.16   \\
Kind  \cite{zhang2019kindling}                  & 19.66   & 0.8519  & 10.42  & 18.06   & 0.8571  & 12.59  & 18.86   & 0.8545  & 11.51   \\
MIRNet \cite{zamir2020learning}                  & 24.14   & \textcolor{blue}{0.8675}  & 7.45   & \textcolor{blue}{28.10}   & \textcolor{blue}{0.9012}  & \textcolor{blue}{5.13}  & \textcolor{blue}{26.12}   & \textcolor{blue}{0.8844}  & \textcolor{blue}{6.29}  \\
SNRNet \cite{xu2022snr}                    & 24.61   & 0.8660   & 6.84   & 21.48   & 0.8717  & 9.15   & 23.05   & 0.8689  & 7.99   \\
Retinexformer    \cite{cai2023retinexformer}       & \textcolor{blue}{25.15}   & \textcolor{blue}{0.8675}  & \textcolor{blue}{6.54}   & 22.79   & 0.8637  & 8.20    & 23.97   & 0.8656  & 7.37    \\ \midrule
TFFormer                & \textcolor{red}{26.13}   & \textcolor{red}{0.8875}  & \textcolor{red}{6.12}   & \textcolor{red}{31.55}   & \textcolor{red}{0.9147}  & \textcolor{red}{3.79}   & \textcolor{red}{28.84}   & \textcolor{red}{0.9011}  & \textcolor{red}{4.95}   \\ \bottomrule
\end{tabular}

\caption{Quantitive comparison between existing SLLIE methods and TFFormer on LOL-V1 and LOL-V2. The proposed TFFormer outperforms the existing methods on existing benchmarking datasets as well. The best and the second-best results are highlighted in \textcolor{red}{red} and \textcolor{blue}{blue}, respectively. }
\label{tab:lolQuant}
\end{table*}

\subsection{Inference Analysis}

Tab. \ref{complexity} demonstrates the inference speed and computational complexity of the proposed TFFormer on our hardware setup. Notably, the proposed TFFormer comprises only 5.87M trainable parameters—substantially fewer than those in well-known transformer models such as Uformer, MIRNet, and SNRNet. Moreover, as a single-stage network, TFFormer enables end-to-end optimization and efficient inference, enhancing performance while significantly reducing computational overhead. It takes just over 0.5 sec to enhance a large dimension low-light input on a mid-end GPU like GTX-3060. It is worth noting we evaluated our method with Float32 precision without performing any optimization. Therefore, the inference speed of TFFormer can be further improved by adopting model compression techniques such as quantization \cite{yang2019quantization}, pruning \cite{li2019deep}, etc., for future usage.

\begin{table}[!htb]
\centering
\rowcolors{1}{gray!15}{white}
\scalebox{0.8}{\begin{tabular}{cccc} \toprule
Input Size  & Param. (M) & Compl. (GMac) & Inf. Time  (ms) \\ \midrule
$256\times256\times3$   &           & 34.52             & 62.21                \\
$512\times512\times3$   &   5.87              & 138.09            & 180.37               \\
$1024\times1024\times3$ &                & 552.35            & 657.78  \\ \bottomrule            
\end{tabular}}

\caption{ Inference analysis of proposed TFFormer.}
\label{complexity}
\end{table}

\section{LSD-TFFormer In Real-world}
\label{extrares}

This section illustrates more results of LSD-TFFormer on diverse scenarios.

\begin{figure*}[!htb]
  \centering
  \includegraphics[width=\linewidth,height=6.7cm]{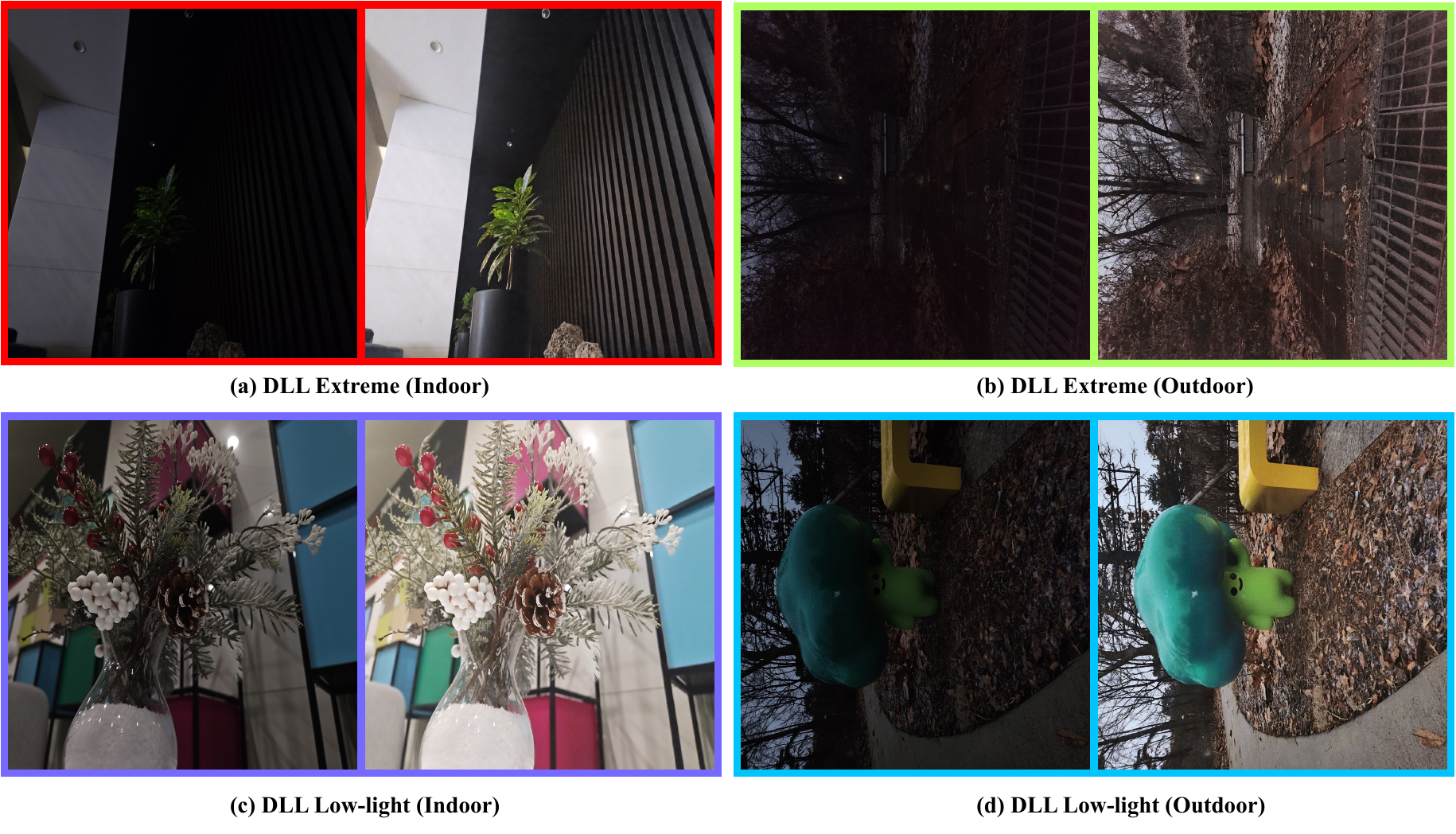}
  \caption{TFFormer performance on DLI subsets, including indoor and outdoor scenes from low-light and extreme lighting conditions. In each pair, on the left is the input image, and on the right is the enhanced image by LSD-TFFormer. }
  \label{fig:dllenhance}
\end{figure*}
\begin{figure*}[!htb]
  \centering
  \includegraphics[width=\linewidth,height=6.7cm]{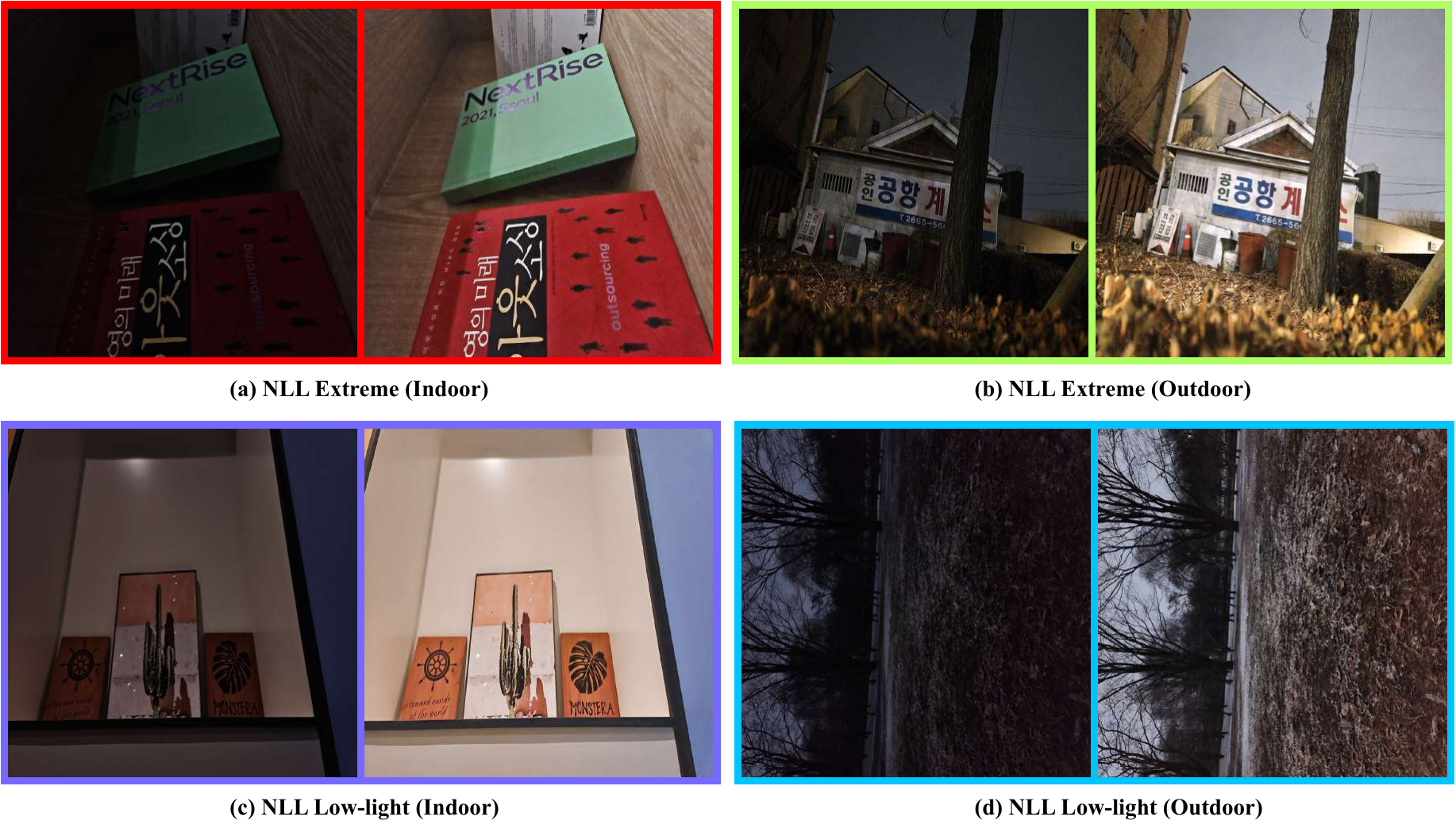}
  \caption{TFFormer performance on NLI subsets, including indoor and outdoor scenes from low-light and extreme lighting conditions. In each pair, on the left is the input image, and on the right is the enhanced image by LSD-TFFormer.}
  \label{fig:nllenhance}
  
\end{figure*}

\subsection{TFFormer on LSD}
Our TFFormer can perform evenly in numerous lighting conditions and scene types. We perform an extensive evaluation of TFFormer in eight subsets of LSD testing set. Fig. \ref{fig:dllenhance} and \ref{fig:nllenhance} illustrate the performance of our TFFormer in all subcategories of the proposed LSD pair benchmark dataset. Please note that the LSD benchmark dataset was collected using various camera sensors of different sizes. These sensors exhibit distinct responses to different lighting conditions. As a result, certain scenes within the LSD testing benchmark dataset may appear either darker or brighter than their actual lighting conditions.

\subsection{TFFormer vs. Existing Method (More Results) }

\begin{figure*}[!htb]
  \centering
  \includegraphics[width=\textwidth, height=5cm]{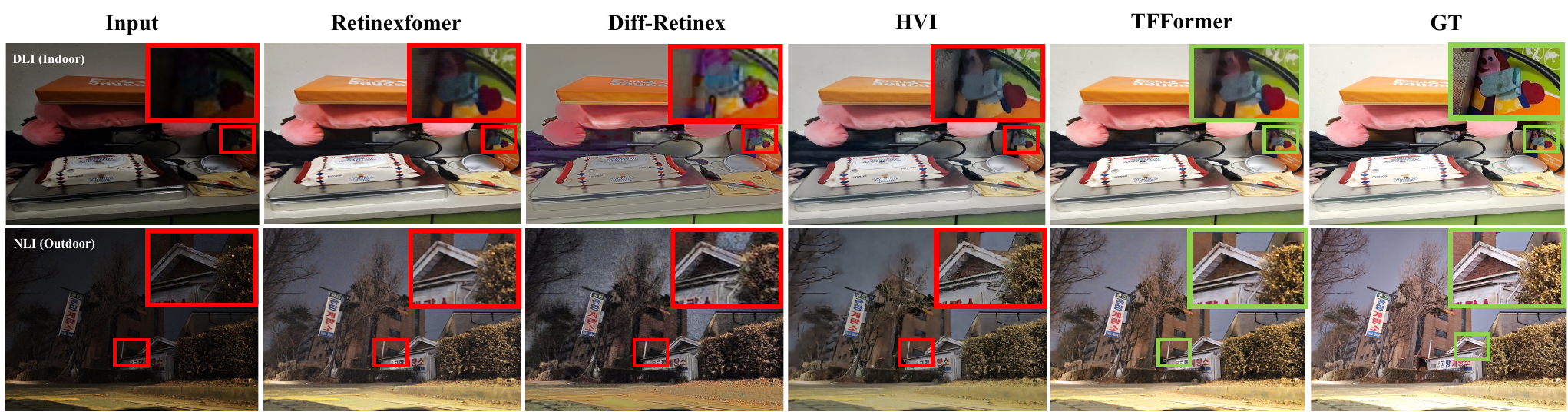}
  \caption{Real-world complex scene enhancement with low-light image enhancement methods. (a) input image capture with Samsung S22 Ultra. (b)-(c) Retinexformer\cite{cai2023retinexformer} with existing datasets. (d)  Retinexformer\cite{cai2023retinexformer} with LSD. (e) Proposed (LSD-TFFormer)}
  \label{fig:visComp1_spp}
\end{figure*}

\begin{figure*}[!htb]
  \centering
  \includegraphics[width=\textwidth]{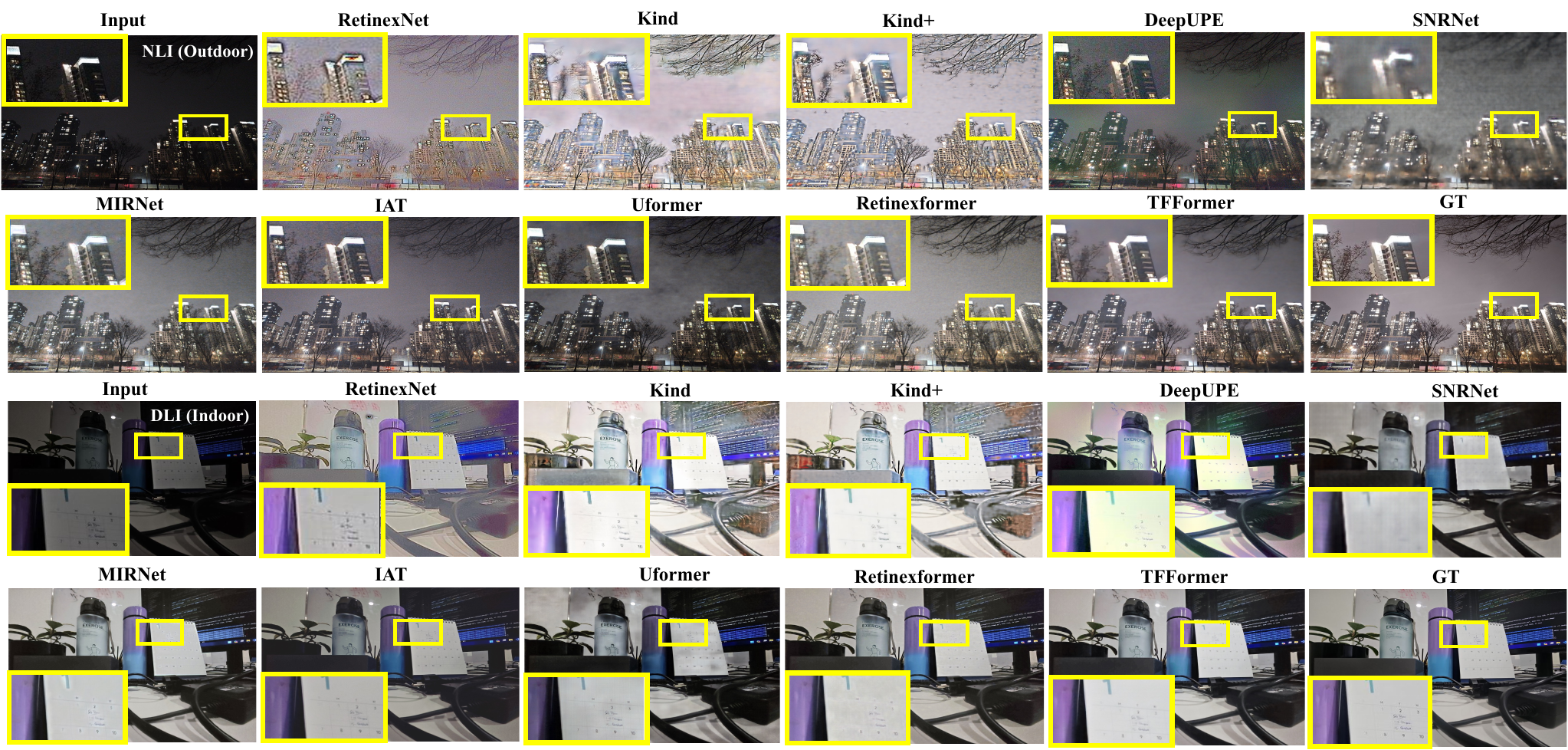}
  \caption{Qualitative comparison of existing SLLIE methods on the LSD dataset. The top scenes show performance on NLI, while
the bottom examples illustrate performance on DLI scene types. TFFormer produces cleaner-plausible images, outperforming existing
methods..}
  \label{fig:visComp2_spp}
\end{figure*}

As we mentioned earlier, the challenges posed by over-enhancement, noise amplification, and color distortion in existing SLLIE methods. We present qualitative comparisons in Figures~\ref{fig:visComp1_spp} and~\ref{fig:visComp2_spp}. These examples underscore TFFormer’s ability to generalize to complex, real-world scenes across varying illumination levels and conditions.

In Figure~\ref{fig:visComp1_spp}, we visually compare TFFormer against three representative SLLIE architectures: Diff-Retinex~\cite{yi2023diff}, HVI~\cite{yan2025hvi}, and RetinexFormer~\cite{cai2023retinexformer}, on both standard low-light (DLI) and challenging noisy low-light (NLI) scenarios. Despite leveraging Retinex priors~\cite{cai2023retinexformer, yi2023diff} or biologically inspired feature fusion~\cite{yan2025hvi}, these baselines frequently suffer from overexposure, structural artifacts, and color distortions—especially under NLI conditions. In contrast, TFFormer consistently delivers results with more natural tone rendering and finer structure preservation, validating the effectiveness of its luminance–chrominance (LC) decoupling and guided refinement mechanism.

Figure~\ref{fig:visComp2_spp} extends the comparison to a broader set of 10 state-of-the-art methods, including classical RetinexNet~\cite{wei2018deep}, Kind(+)\cite{zhang2019kindling, zhang2021beyond}, DeepUPE\cite{wang2019underexposed}, MIRNet~\cite{zamir2020learning}, IAT~\cite{cui2205you}, Uformer~\cite{wang2022uformer}, SNRNet~\cite{xu2022snr}, GSAD~\cite{hou2023global}, and LYT~\cite{brateanu2025lyt}. Across both DLI and NLI cases, TFFormer produces perceptually coherent, cleaner outputs with better structure and fewer color artifacts. Prior RGB-based transformer and CNN methods often suffer from over-smoothing (e.g., MIRNet, Uformer) or produce strong unnatural color casts (e.g., DeepUPE, SNRNet). Retinex-based designs, despite strong PSNR, tend to yield inconsistent tones under challenging lighting, likely due to excessive or misaligned enhancement. These observations, supported by both qualitative and quantitative evidence, underscore the robustness and generalization capacity of TFFormer in complex real-world scenes.

Together, these visual results reinforce the paper’s central hypothesis: existing SLLIE models struggle to preserve structure and realism under challenging real-world scenarios, while TFFormer, trained on the proposed LSD dataset and guided by LC-aware modules, produces high-fidelity enhancements with superior generalization.

\begin{figure*}[!htb]
  \centering
  \includegraphics[width=\textwidth, height=5cm]{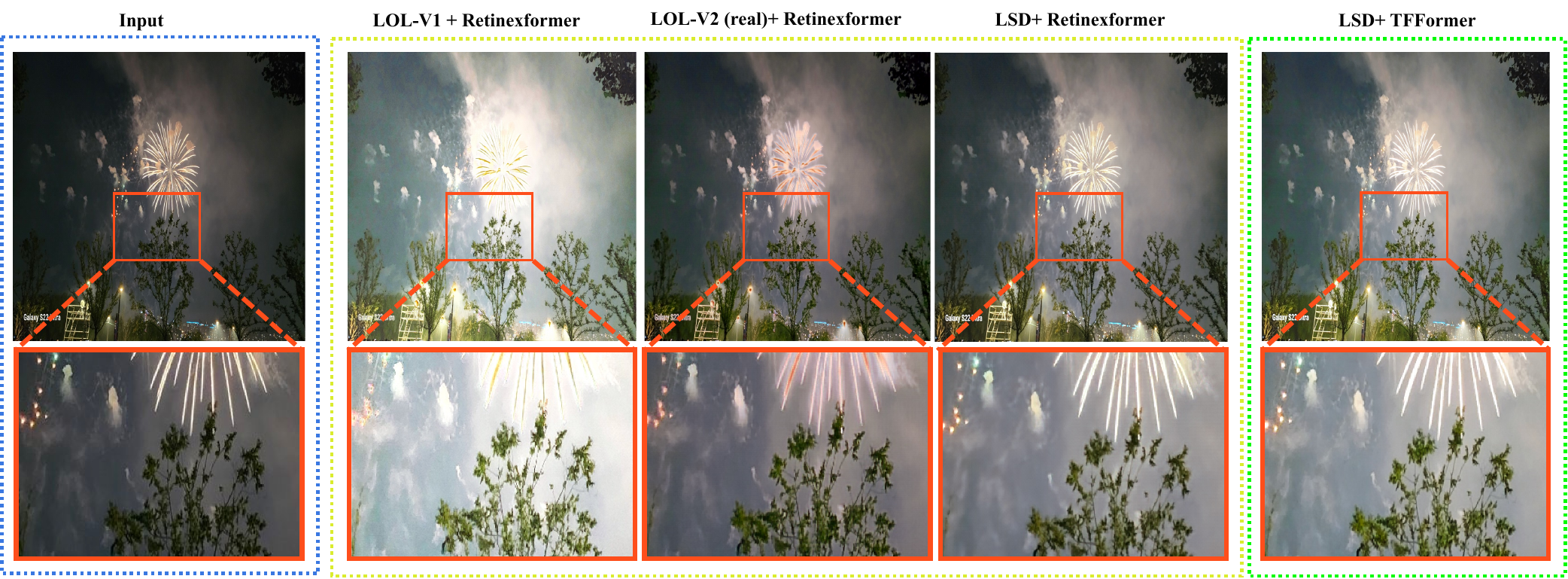}
  \caption{Real-world complex scene enhancement with low-light image enhancement methods. (a) input image capture with Samsung S22 Ultra. (b)-(c) Retinexformer\cite{cai2023retinexformer} with existing datasets. (d)  Retinexformer\cite{cai2023retinexformer} with LSD. (e) Proposed (LSD-TFFormer)}
  \label{fig:losdonsota}
\end{figure*}

\subsection{LSD on Deep SLLIE Methods}

The primary motivation of this study is to advance SLLIE methods by providing a reliable dataset for training deep-learning models on real-world scenarios and benchmarking their performance across diverse conditions. To evaluate this, we trained the state-of-the-art Retinexformer \cite{cai2023retinexformer} on existing LOL-V1, LOL-V2, and our proposed LSD dataset. Subsequently, we tested the model on complex scenes out of this dataset. Fig. \ref{fig:losdonsota} illustrates the performance of Retinexformer in enhancing these real-world complex scenes.

Notably,  Retinexformer is one of the top-performing methods on LOL-V1 and LOL-V2. However, we observed that even its pretrained model (obtained directly from the official repository) often produces severe artifacts and overexposed regions in challenging scenarios trained with existing SLLIE datasets. In contrast, training Retinexformer on the proposed LSD dataset enables it to generate more natural and visually appealing results. Furthermore, our TFFormer surpasses its counterparts by producing cleaner and more plausible images, thanks to its LC Encoding and LC Guidance mechanisms. These results underline the significant contributions of the LSD dataset and TFFormer in tackling real-world SLLIE challenges and their practicability in improving SLLIE methods.

\begin{figure*}[!htb]
  \centering
  \includegraphics[width=\textwidth, height=15.5cm]{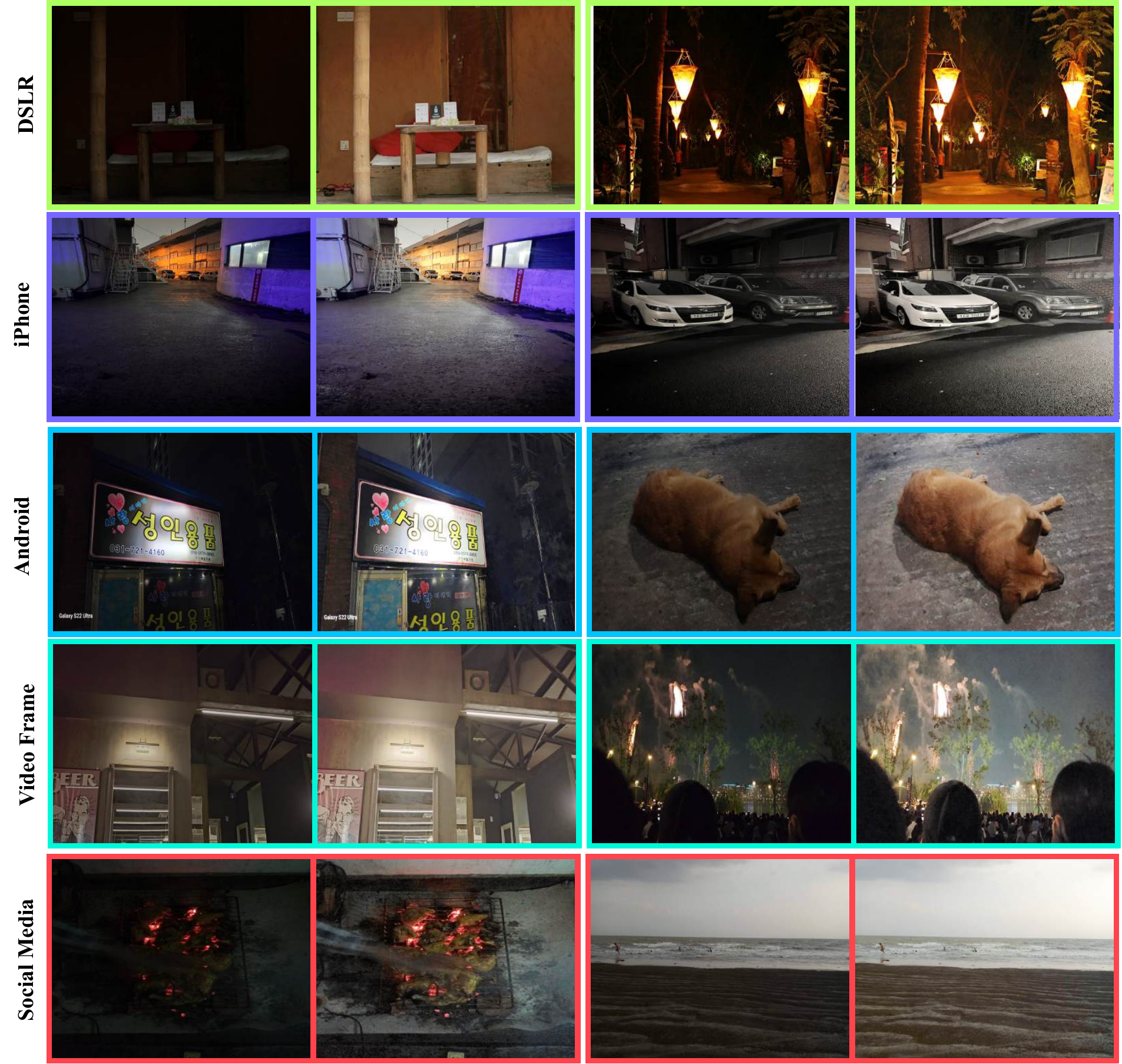}
  \caption{Few more examples of generic SLLIE obtained by LSD-TFFormer. In each pair, on the left is the input image, and on the right is the enhanced image by LSD-TFFormer.}
  \label{fig:beyondsupp}
\end{figure*}

\subsection{LSD-TFFormer on Real-world Scenes}
\textbf{Visual Comparison.} Fig. \ref{fig:beyondsupp} illustrates more examples of real-world SLLIE obtained by the proposed LSD-TFFormer. Our method can handle diverse scenes obtained by numerous hardware and lighting conditions. Our LSD-TFFormer is also effective in enhancing images stored on social media (e.g., Facebook, Twitter, Instagram, etc.). These popular social platforms contain billions of user images captured with old camera hardware in low-light conditions. Our proposed method opens a new life to these images by enhancing and providing them with a modern touch.

\textbf{User Study.} We also performed a separate user study to verify the mass acceptance of our LSD-TFFormer in real-world scenarios. In this study, 35 participants aged 15 to 62 evaluated enhanced outputs across various lighting conditions and capture sources. Each participant was shown three randomly selected low-light versus enhanced image pairs from multiple SLLIE datasets and asked to choose the image they preferred, based solely on aesthetic appeal and without knowledge of the study's purpose.

As summarized in Table~\ref{tab:userstudy}, over 82\% of participants preferred the outputs generated by LSD-TFFormer. The preference was particularly strong in categories involving older social media photos and noisy video frames, where traditional models tend to fail due to compression artifacts and low signal-to-noise ratios. This preference distribution supports the claim that our model generalizes beyond controlled lab settings and remains robust across highly diverse, real-world image sources.

\begin{figure*}[!htb]
  \centering
  \includegraphics[width=\textwidth, height=5cm]{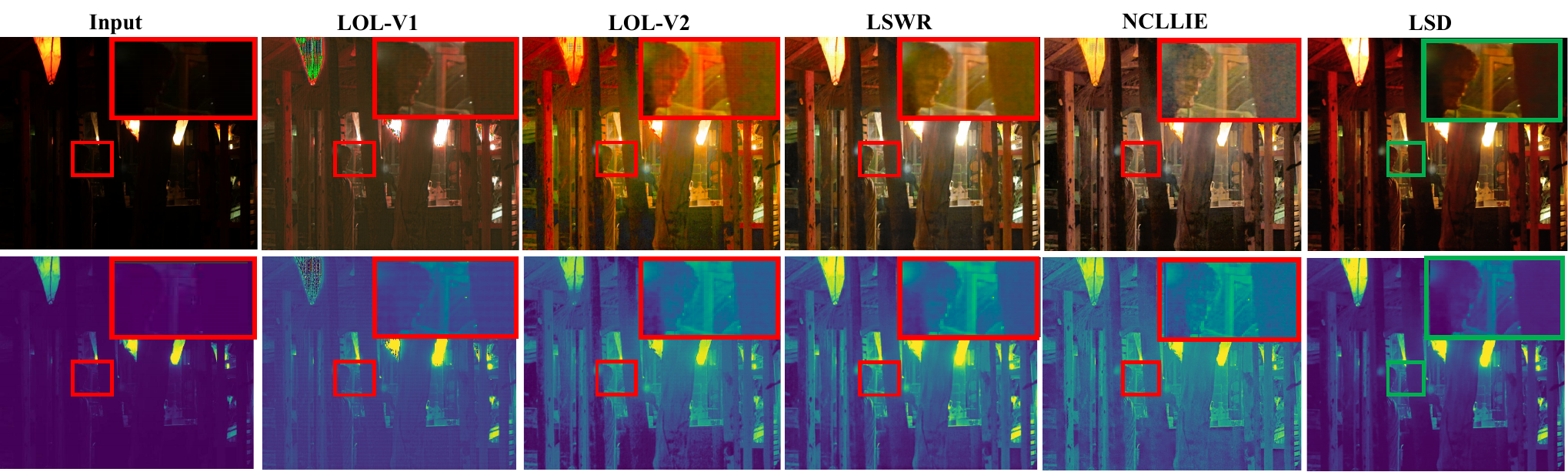}
  \caption{Real-world complex scene enhancement with low-light image enhancement methods. (a) input image capture with Samsung S22 Ultra. (b)-(c) Retinexformer\cite{cai2023retinexformer} with existing datasets. (d)  Retinexformer\cite{cai2023retinexformer} with LSD. (e) Proposed (LSD-TFFormer)}
  \label{fig:lsdvssllied_supp}
\end{figure*}

Fig.~\ref{fig:lsdvssllied_supp} visually reinforces these findings by comparing LSD-trained models against those trained on existing datasets, such as LOL-V1/V2, LSRW, and NCLLIE. While models trained on traditional datasets often produce flat or over-smoothed results, LSD-trained TFFormer outputs exhibit more natural contrast, reduced artifacts, and better color fidelity. In many cases, the LSD-enhanced images preserved ambient lighting cues and semantic details that were lost in other reconstructions. This aligns with user preferences and demonstrates the practical utility of our dataset and model for consumer-grade image enhancement applications.

\begin{table}[!htb]
    \centering
    \rowcolors{1}{gray!15}{white}
    \scalebox{0.95}{\begin{tabular}{ccc}
    \toprule
    {Device}  & {Low-light $\uparrow$} & {LSD-TFFormer $\uparrow$} \\ \midrule
    DSLR           & 0.0952    & 0.9048          \\
    Video (Frames) & 0.3095    & 0.6905          \\
    iPhone         & 0.1905    & 0.8095          \\
    Android        & 0.2063    & 0.7937          \\
    Social Media   & 0.0714    & 0.9286          \\ \midrule
    Average        & 0.1746    & \textcolor{red}{0.8254}     \\ \bottomrule
    \end{tabular}}
    \caption{User study on LSD-TFFormer. In 82\% of cases, user preferred LSD-TFFormer over low-light images.}
    \label{tab:userstudy}
\end{table}

\subsection{Visual Improvement on Vision Task}

\begin{figure*}[!htb]
  \centering
  \includegraphics[width=\textwidth]{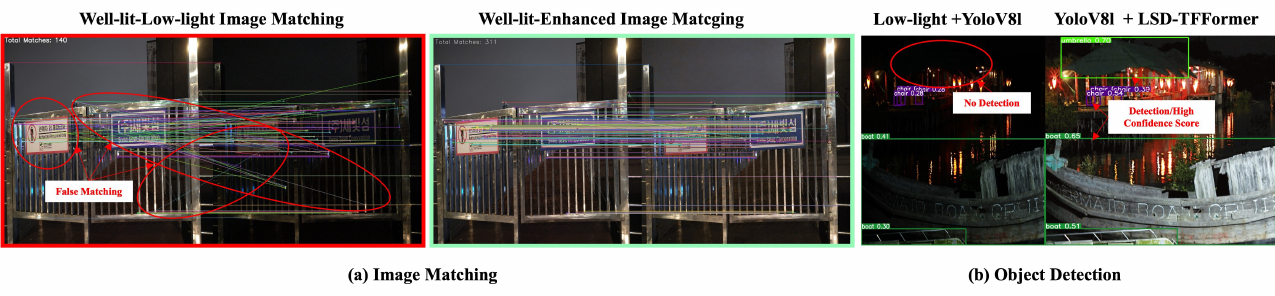}
  \caption{LSD-TFFormer usability in vision tasks. (a) Image matching. (b) Object Detection}
  \label{fig:visiontask}
\end{figure*}

Apart from aesthetical use cases, the proposed method can also accelerate everyday vision tasks. Fig. \ref{fig:visiontask} Illustrates the performance gain achieved in two prominent vision tasks by incorporating our proposed method. Our approach significantly enhances the performance of widely utilized object detection (OD) \cite{ultralytics} algorithms and facilitates improved matching of images \cite{rublee2011oRB} in low-light conditions. Notably, matching and registering low-light and well-lit images to make a paired dataset for SLLIE is extremely difficult. However, our method can support future studies by seamlessly matching collected low-light and well-lit images to enrich SLLIE research. 

\subsection{LSD-TFFormer vs Night mode}
\label{nightmode}
Night-mode photos are typically produced by combining multiple shots with different exposure settings. Comparing multi-shot approaches like night mode with single-shot methods such as LSD-TFFormer is unfair. However, to push the limit and study the feasibility of single-shot SLLIE in a broader aspect, we evaluated and compared our LSD-TFFormer with night mode photos. Thus, we mounter our smartphone on a fixed point and fixed its focal point. Later, we captured one scene in auto mode and another by enabling the dedicated night mode.

\begin{figure}[!htb]
  \centering
  \includegraphics[width=\linewidth, height=4cm]{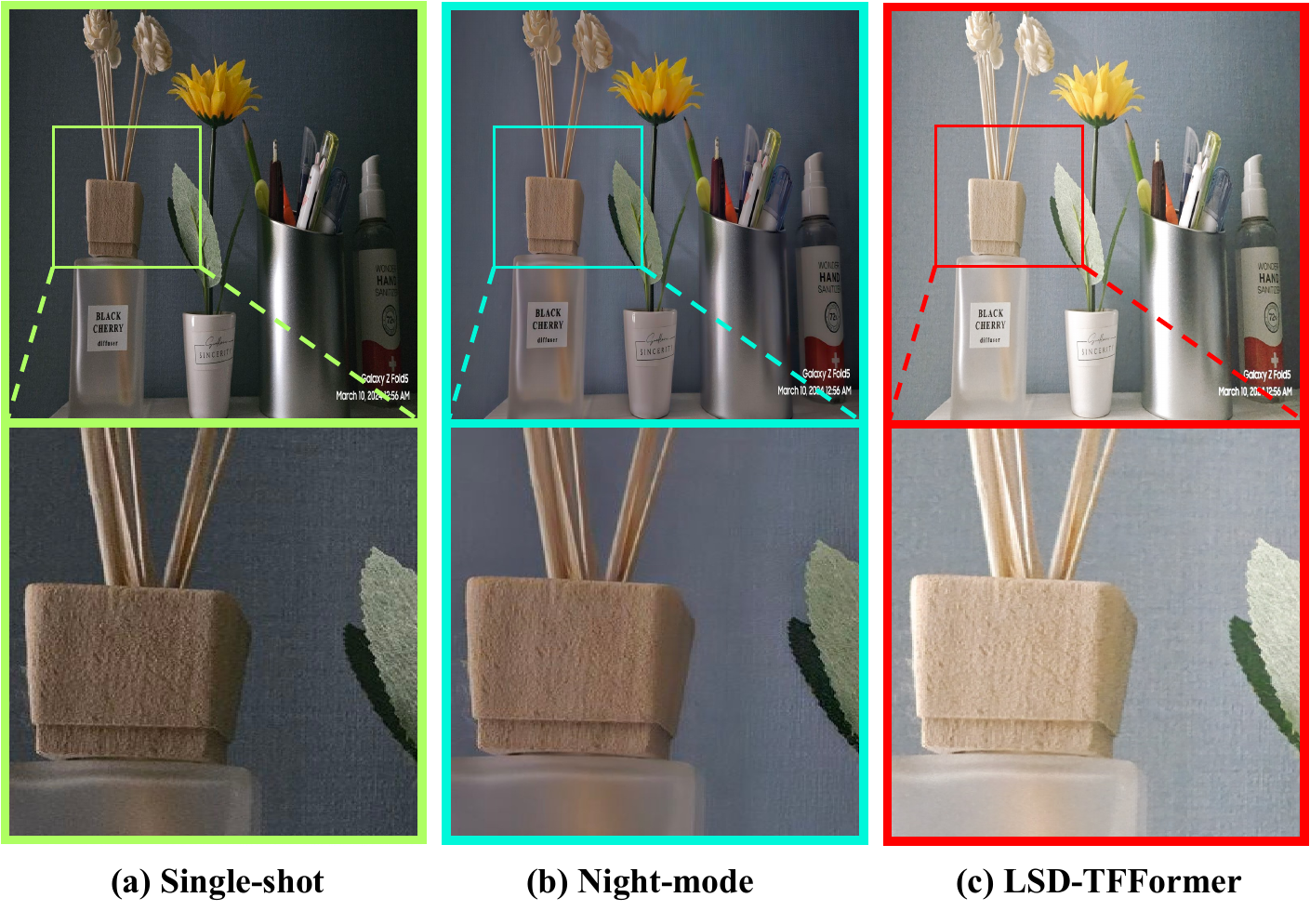}
  \caption{Comparison between LSD-TFFormer and night mode. Scenes capture under 20 lux with Samsung Galaxy Z fold 5. (a) Z Fold 5 without night mode. (b) Z Fold 5 with night mode. (c) Z Fold 5 without night mode scene enhanced by LSD-TFFormer.}
  \label{fig:lsdvsnight}
\end{figure}

We enhanced the auto-mode photo (without night mode) with our LSD-TFFormer. Fig. \ref{fig:lsdvsnight} illustrates the comparison between LSD-TFFormer and night mode from Samsung Galaxy Z fold 5. We found that such dedicated night photography mode of smartphones is a trend to produce smooth images that lack salient details. On the other hand, our method can produce brighter images compared to the multi-shot processing while maintaining the salient details of the short-exposure images.

\subsection{Failure Case}
\label{failure}
\begin{figure}[!htb]
  \centering
  \includegraphics[width=\linewidth, height=4cm]{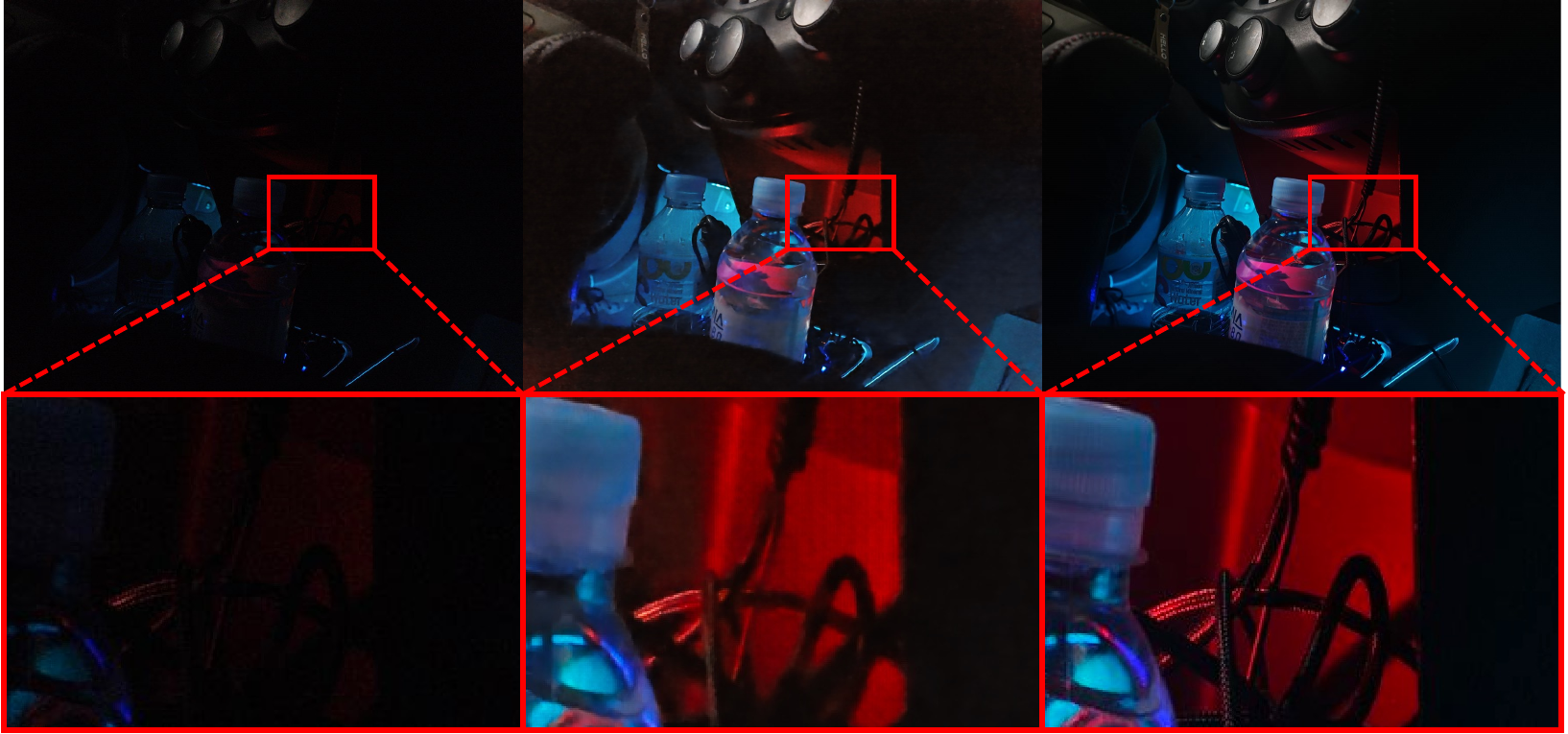}
  \caption{ Example of failure case. For NLL scenes, TFFormer can illustrate visible noise in some extreme cases (e.g., under 1 lux). Left: NLL input (under 1 lux ), right: LSD-TFFormer. }
  \label{fig:failure}
\end{figure}

Despite promising results in numerous test cases, our TFFormer can produce visible noisy regions in NLL scenes captured under 1 lux. Fig. \ref{fig:failure} illustrates such an example of a failure case. We planned to address such extremely challenging cases in future studies.

